\PassOptionsToPackage{prologue,dvipsnames}{xcolor}

\documentclass[letterpaper]{article} 
\usepackage{template}  
\usepackage{times}  
\usepackage{helvet} 
\usepackage{courier} 
\usepackage[hyphens]{url} 
\usepackage{graphicx} 
\usepackage{natbib} 
\usepackage{caption} 
\usepackage{algorithm}
\usepackage{algorithmic}
\usepackage{amsmath}
\usepackage{booktabs}
\usepackage{tikz}
\usepackage{comment}
\usepackage{hyperref}
\usepackage{amsmath,amssymb,amsfonts}
\usepackage[dvipsnames]{xcolor} 

\newcommand{\up}{\textcolor{ForestGreen}{$\blacktriangle$}}
\newcommand{\down}{\textcolor{BrickRed}{$\blacktriangledown$}}
\newcommand{\upworse}{\textcolor{BrickRed}{$\blacktriangle$}}
\newcommand{\downbetter}{\textcolor{ForestGreen}{$\blacktriangledown$}}
\newcommand{\tsp}[1]{${}^{\text{#1}}$}

\usepackage{newfloat}
\usepackage{listings}

\floatstyle{ruled}
\newfloat{listing}{tb}{lst}{}
\floatname{listing}{Listing}

\title{Neural Reasoning Networks: Efficient Interpretable Neural Networks With Automatic Textual Explanations}
\author{%
Stephen Carrow$^{1}$ \quad Kyle Harper Erwin$^1$ \quad Olga Vilenskaia$^1$ \quad Parikshit Ram$^1$ \\
\textbf{Tim Klinger}$^1$ \quad \textbf{Naweed Aghmad Khan}$^1$ \quad \textbf{Ndivhuwo Makondo}$^1$ \quad \textbf{Alexander Gray}$^2$  \\
}
\affiliations {
    \textsuperscript{\rm 1}IBM, 
    \textsuperscript{\rm 2}Centaur AI Institutue\\
    \{steve.carrow, kyle.erwin, olga.vilenskaia, parikshit.ram, naweed.khan, ndivhuwo.makondo\}@ibm.com,\\ 
    tklinger@us.ibm.com, skirmilitor@gmail.com
}

\begin{document}

\maketitle

\begin{abstract}

Recent advances in machine learning have led to a surge in adoption of neural networks for various tasks, but lack of interpretability remains an issue for many others in which an understanding of the features influencing the prediction is necessary to ensure fairness, safety, and legal compliance. In this paper we consider one class of such tasks, tabular dataset classification, and propose a novel neuro-symbolic architecture, Neural Reasoning Networks (NRN), that is scalable and generates logically sound textual explanations for its predictions. NRNs are connected layers of logical neurons which implement a form of real valued logic. A training algorithm (R-NRN) learns the weights of the network as usual using gradient descent optimization with backprop, but also learns the network structure itself using a bandit-based optimization. Both are implemented in an extension to PyTorch (\href{https://github.com/IBM/torchlogic}{available on GitHub}) that takes full advantage of GPU scaling and batched training. Evaluation on a diverse set of 22 open-source datasets for tabular classification demonstrates performance (measured by ROC AUC) which improves over multi-layer perceptron (MLP) and is statistically similar to other state-of-the-art approaches such as Random Forest, XGBoost and Gradient Boosted Trees, while offering 43\% faster training and a more than 2 orders of magnitude reduction in the number of parameters required, on average. Furthermore, R-NRN explanations are shorter than the compared approaches while producing more accurate feature importance scores.

\end{abstract}

\section{Introduction}

Classifying tabular data has long been a fundamental task in ML/AI and recent advances in AI have led to increased adoption in various industries, but the complexity of these systems has made them black boxes that are difficult to understand and interpret \cite{adadiPeekingBlackBoxSurvey2018a,linardatosExplainableAIReview2021a,barredoarrietaExplainableArtificialIntelligence2020}. This lack of transparency is a major obstacle to the adoption of AI in sensitive domains such as healthcare, and increasingly practitioners and researchers aim to explain black-box model predictions \cite{rudinStopExplainingBlack2019a}. A variety of tools have been designed towards that end \cite{guidottiSurveyMethodsExplaining2018}, however popular methods such as SHAP \cite{DBLP:journals/corr/LundbergL17} and LIME \cite{DBLP:journals/corr/RibeiroSG16} have been shown to be susceptible to adversarial attacks \cite{yuanSimpleScoringFunction2023,slackFoolingLIMESHAP2020} and fail to effectively explain neural networks and generalized additive models (GAMs) \cite{carmichaelHowWellFeatureAdditive2023}.  As a result, these post-hoc explainers can produce misleading explanations, especially when users misunderstand or over trust them. Recent work even considers the use of black-box models to explain other black-box models, such as with a pre-trained LLM \cite{kroeger2023are} or Diffusion model \cite{madaan2023diffusionguided}, further obfuscating the explanation generation process.

One potential solution to address these drawbacks is to use inherently interpretable models, which can produce their own explanations that  faithfully represent the model's computation \cite{rudinStopExplainingBlack2019a}.  This approach is explored in, for example, Neural Additive Models (NAM) \cite{agarwal2021neural}, and CoFrNet (CFN) \cite{puriCoFrNetsInterpretableNeural2021}, which have performance that is statistically similar to deep-tree-ensemble-based methods that are known to perform well for tabular classification. 

\tikzset{%
  every neuron/.style={
    circle,
    draw,
    minimum size=0.75cm
  },
  neuron missing/.style={
    draw=none, 
    scale=2,
    text height=0.3cm,
    execute at begin node=\color{black}$\vdots$
  },
}

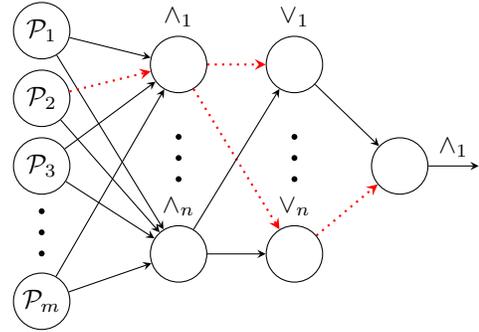
\begin{figure}
  \begin{center}
    \begin{tikzpicture}[x=1.4cm, y=0.9cm, >=stealth]
    
    \foreach \m/\l [count=\y] in {1,2,3,missing,4}
     \node [every neuron/.try, neuron \m/.try] (input-\m) at (0,2.5-\y) {};
    
    \foreach \m [count=\y] in {1,missing,2}
      \node [every neuron/.try, neuron \m/.try ] (conjunction-\m) at (1.3,2.4-\y*1.4) {};
    
    \foreach \m [count=\y] in {1, missing,2}
      \node [every neuron/.try, neuron \m/.try ] (disjunction-\m) at (2.4,2.4-\y*1.4) {};
    
    \foreach \m [count=\y] in {1}
      \node [every neuron/.try, neuron \m/.try ] (output-\m) at (3.4,0.5-\y) {};
    
    \foreach \l [count=\i] in {1,n}
      \node [above] at (conjunction-\i.north) {$\wedge_\l$};
    
    \foreach \l [count=\i] in {1,n}
      \node [above] at (disjunction-\i.north) {$\vee_\l$};
    
    \foreach \l [count=\i] in {1}
      \draw [->] (output-\i) -- ++(0.75,0)
        node [above, midway] {$\wedge_\l$};
    
    \foreach \i in {1,...,4}
      \foreach \j in {1,...,2}
        \ifthenelse{\i=2 \and \j=1}{\draw [dotted, red,thick, ->] (input-\i) -- (conjunction-\j)}{\draw [->] (input-\i) -- (conjunction-\j)};
    
    \foreach \i in {1,...,2}
      \foreach \j in {1,...,2}
        \ifthenelse{\i=1}{\draw [dotted, red, thick, ->] (conjunction-\i) -- (disjunction-\j)}{\draw [->] (conjunction-\i) -- (disjunction-\j)};
    
    \foreach \i in {1,...,2}
      \foreach \j in {1}
        \ifthenelse{\i=1}{\draw [->] (disjunction-\i) -- (output-\j)}{\draw [dotted, red, thick, ->] (disjunction-\i) -- (output-\j)};
        
    \foreach \l [count=\x from 0] in {1}
    \node [align=center, above] at (\x*2,2) {};
    \node (nx1) at (input-1) {$\mathcal{P}_1$};
    \node (nx2) at (input-2) {$\mathcal{P}_2$};
    \node (nx3) at (input-3) {$\mathcal{P}_3$};
    \node (nx4) at (input-4) {$\mathcal{P}_m$};
    
    \end{tikzpicture}
    
  \end{center}
  \caption{A two layer NRN for R-NRN (Figure~\ref{fig:BRNAlgoVisual}  and Algorithm~\ref{Algo:BRN}), with dropped edges in red.  The resulting network is initialized with sparse connections.}
    \label{Fig:BRNNeuralReasoningNetwork}
\end{figure}

However, inherently interpretable models are not all equal in their explanation quality, predictive accuracy, or practical utility such as training speed or parameter counts. For example, our experiments show that the best NAM and CFN models after hyper-parameter tuning have approximately 861K and 37K parameters respectively. In this paper, we introduce the NRN architecture and the R-NRN supervised classification algorithm.  NRNs use Weighted Lukasiewicz Logic, which forms the foundation of interpreting the resulting network after training.  
Each node is interpreted as either an ``And`` or ``Or" operation, so the model itself is composed of logical operations over the input features.  R-NRN, is constructed with sparsely connected logical nodes, as depicted in Figure~\ref{Fig:BRNNeuralReasoningNetwork}, requires only 1K parameters and trains, on average, 43\%\footnote{Avg. R-NRN trial time versus the mean of avg. trial times for NAM and CFN in Table~\ref{Table:ExperimentalResults} \label{foot:speed}} faster compared NAM and CFN, while its performance in terms of ROC AUC is statistically similar to RF, XGB, and GBT as summarized in Table~\ref{Table:ExperimentalResults}.

Furthermore, while methods like NAM and CFN are interpretable, explanations take the form of directional feature importance showing only relative contribution from features, in contrast to our approach that produces a natural language description of the logical rule set for a prediction. This distinction is important in that the rule set produced contains not only direction of influence and relative importance for each rule, but also the specific conditions on a sample that must be satisfied, such as relational constraints of the form  $f_1 > X$, for $f_1$ a real-valued feature, and $X$ a learned boundary value.  In addition, explanation sizes for methods based on feature importance, including NAM, CFN and even those produced with SHAP, increase with the number of features in the data set while R-NRN explanation size is proportional to the size of the learned rules, which may have much fewer features. We show experimentally that R-NRN training results in NRN model explanations that are 31\% smaller on average than these feature-importance-based methods for sample level explanations.

While the NRN learned using R-NRN may be rather complex,  we achieve concise sample level explanations by simplifying verbose logic with straight forward rules, such as $\textit{AND}(x > a; x > b) = x > \max(a, b)$, and we can ignore all disjunction branches which are not true for a sample leaving only the logic where all conditions are met. Since the sample under analysis will satisfy all the logic that remains following such transformations, the explanation can be simplified to a single conjunction. Details of explanation generation are discussed in \textbf{Section 3}.

In summary, we contribute the following in this work:

\begin{enumerate}
    \itemsep0em
    \item \textbf{Neural Reasoning Networks}:  A neuro-symbolic AI architecture that uses a Modified Weighted Lukasiewicz Logic constructing interpretable networks that leverage AutoGrad, batched training, and GPU acceleration.
    \item \textbf{Bandit Reinforced Neural Reasoning Network (R-NRN)}: A \textit{supervised classification} algorithm for \textit{tabular data} with predictive performance that is statistically similar to RF, XGB, and GBT and which produces compact sample level explanations of logical rules sets.
    \item \textbf{Automatic explanation generation}:  A novel algorithm that leverages the structure of NRNs, algebraic manipulations of Weighted Lukasiewicz Logic, and logical interpretation of NRN nodes to produce natural language explanations of the models' predictions. 
    \item \textbf{Rigorous evaluation}: An elaborate evaluation with various models and tabular datasets, highlighting the competitiveness of our proposed R-NRN. Relevant to the research on neural networks for tabular data, our results show that R-NRN, NAM, CFN, and DAN outperform various tabular NN with a significant margin and are on par with RF, XGB, and GBT, contrary to existing results.
\end{enumerate}

The remainder of the paper presents the details of NRNs and R-NRN along with our experimental results and analysis of our algorithm. In \textbf{Section 2}, we review the related work in explainable machine learning and discuss limitations of the previously published techniques. In \textbf{Section 3}, we introduce Neural Reasoning Networks, detail the R-NRN algorithm for classification, and our explanation algorithm. In \textbf{Section 4}, we share the details of an extensive empirical study comparing the predictive performance of R-NRN to existing algorithms. In addition, we analyze R-NRN explanation quality. In \textbf{Section 5} we discuss limitations and conclude our work with future directions for this line of research.

\section{Related work}

In this section we review related works and discuss how they are positioned within the landscape of AI/ML algorithms used for supervised tabular classification, as well as their relation to explainable AI (XAI).

We begin with post-hoc explanation of a trained AI/ML model, as this is a highly common \cite{rudinStopExplainingBlack2019a} and flexible approach to XAI in practice. Application of post-hoc explainers like SHAP and LIME are traditionally used to explain \textit{non-interpretable models}. Well-known instances of non-interpretable models are Random Forests (RF) \cite{breiman2001random}, Gradient Boosted Trees (GBT) \cite{10.1214/aos/1013203451}, eXtreme Gradient Boosting (XGB) \cite{DBLP:journals/corr/ChenG16}, and Multilayer perceptron (MLP). More recently, research has focused on development of neural-methods for tabular data leveraging modern architectures like Transformer. A recent example is the FTTransformer (FTT) \cite{gorishniyRevisitingDeepLearning2021}. FTT is a simple adaptation of the Transformer architecture for tabular data, transforming all features (categorical and numerical) to embeddings and applying a stack of Transformer layers to the embeddings. FTT outperforms other deep learning methods according to the authors. 

\citeauthor{chenDANetsDeepAbstract2022} proposed another non-interpretable neural method for learning from tabular data.  They contribute a flexible neural component for tabular data called the Abstract Layer, which learned to explicitly group correlative input features and generate higher-level features for semantics abstraction \cite{chenDANetsDeepAbstract2022}. This approach is somewhat similar to self-attention in that it learns sparse feature selections.  The authors also proposed Deep Abstract Networks (DANets) which use these Abstract Layers. The authors show that DANets are effective and have superior computational complexity compared to competitive methods.

Many more neural methods for learning from tabular data exist.  An extensive review of all such literature is outside the scope of this paper and other works such as \cite{9998482} provide such surveys. 

\begin{table*}[!ht]
\centering
\begin{tabular}{lp{12.9mm}|p{6.9mm}p{5.5mm}p{5.5mm}p{5.5mm}p{5.5mm}p{5.5mm}p{5.5mm}p{5.5mm}p{5.5mm}p{5.5mm}}
\toprule
                 & R-NRN          & BRCG         & DIF          & NAM             & CFN          & FTT              & DAN             & MLP             & RF     & XGB         & GBT \\ \midrule
Logical          & \textbf{Yes} & \textbf{Yes} & \textbf{Yes} & No              & No           & No               & No              & No              & No     & No          & No \\    
Linear Scoring   & \textbf{Yes} & No$^*$       & No$^*$       & \textbf{Yes}    & \textbf{Yes} & No$^*$           & No$^*$          & No$^*$          & No$^*$ & No$^*$      & No$^*$ \\      
Case Reasoning   & \textbf{Yes} & \textbf{Yes} & No           & No              & No           & No               & No              & No              & No     & No          & No \\
GPU Scaling      & \textbf{Yes} & No           & \textbf{Yes} & \textbf{Yes}    & \textbf{Yes} & \textbf{Yes}     & \textbf{Yes}    & \textbf{Yes}    & No     & No          & No \\
Uses AutoGrad     & \textbf{Yes} & No           & \textbf{Yes} & \textbf{Yes}    & \textbf{Yes} & \textbf{Yes}    & \textbf{Yes}    & \textbf{Yes}    & No     & No          & No \\
Batched Training & \textbf{Yes} & No           & \textbf{Yes} & \textbf{Yes}    & \textbf{Yes} & \textbf{Yes}     & \textbf{Yes}    & \textbf{Yes}    & No     & No          & No \\\bottomrule    
\end{tabular}
\caption{
Assessment of algorithms against \cite{rudinStopExplainingBlack2019a} challenges for intepretability and assessment of scalability attributes AutoGrad and Batched Training.  Methods with $\textit{No}^*$ listed for linear scoring can be estimated using SHAP.}
\label{tab:RudinResults}
\end{table*} 

Another branch of research focuses on \textit{inherently interpretable models}.  These methods should themselves produce faithful explanations representing the model's computation \cite{rudinStopExplainingBlack2019a}. Such models produce explanations and allow for interpretation in a variety of ways.  GAMs, one such class of interpretable models, are a type of statistical model that extends the concept of generalized linear models (GLMs) by allowing for more flexible, non-parametric relationships between the target variable and the predictor variables (features) \cite{hastie2017generalized}. Formally, GAMs are defined as 
\begin{equation}\label{Equation:GAM}
    g(\mathbb{E}[y])=\beta+f_1(x_1)+f_2(x_2)+\cdots+f_K(x_K)
\end{equation}
where $\mathbf{x}$ is the input vector with $K$ features, represented as $(x_1, x_2, ..., x_K)$, $y$ is the target variable, $g$ represents the link function, and  each $f_i$ is a smoothing function.  Models in this class aim to produce explanations using the weights assigned to the linear combination. In the case of Neural Additive Models (NAMs), this is a combination of neural networks each focused on a single feature.  NAM can also be explained at a global level by visualizing the shape functions learned by each neural network \cite{agarwal2021neural}. 

Scalable Interpretability via Polynomials (SPAM) \cite{dubey2022scalable} are similar to NAM, except that they efficiently model higher order feature interactions with polynomials.  While the authors show that this approach does improve AUC and RMSE on a variety of datasets, they only evaluate interpretability for the SPAM-linear that does not model those higher level feature interactions.  Furthermore, the authors point out that such polynomial feature interactions are less interpretable. 

Finally, Puri \textit{et al}. proposed a new neural network architecture called Continued Fraction Network (CFN) which was inspired by continued fractions from number theory \cite{puriCoFrNetsInterpretableNeural2021}.  This method is another approach to producing models with interpretable weights.  The authors show that CFNs are efficient learners and because they represent linear combinations of features they can be interpreted in a similar manner to NAM, SPAM, and other GLMs.

\textit{Logic or rule based models} are another common approach to creating interpretable models.  These methods produce a rule set that describes the positive class for tabular classification problems.  For example, the Boolean Rule Column Generation (BRCG) approach learns interpretable Boolean rules in disjunctive normal form (DNF) or conjunctive normal form (CNF) for classification \cite{dashBooleanDecisionRules2018}. It uses column generation to efficiently search over an exponential number of candidate clauses without pre-mining or restrictions.  The resulting model can be interpreted by examining the rule set induced during training.  

Rule based models have also been combined with linear scoring, meaning scores or weights are assigned to rules or features based on
their impact within in the model. For instance, \citeauthor{weiGeneralizedLinearRule2019} proposed a new approach to building generalized linear models using rule-based features, also known as rule ensembles, for regression and probabilistic classification. The approach uses the same binary column generation technique as BRCG. The method is shown to obtain better accuracy-complexity trade-offs than existing rule ensemble algorithms and is competitive with less interpretable benchmark models \cite{weiGeneralizedLinearRule2019}. One interpreting the model can understand both the rules that lead to a prediction and the relative influence of those rules.

Another example of combining logic and weights is Logical Neural Networks (LNN) \cite{DBLP:journals/corr/abs-2006-13155}, a framework that combines neural networks and Weighted Lukasiewicz Logic to score logical nodes with a linear weight.  The framework can minimize logical contradiction, enabling it to handle inconsistent knowledge and make open-world assumptions.  This approach is highly related to our work, however unlike our work, the authors do not propose a \textit{supervised classification} method to solve traditional ML problems on tabular data, and leverage First Order Logic rather than the Propositional Logic of NRNs.

Not all logic or rules based models maintain interpretability, however.  Deep Differentiable Logic Gate Networks (DIF), for example, learn a real-valued representation of logic gates that can then be discretized to a traditional output logic gate network for fast inference \cite{NEURIPS2022_0d3496dd}. DIF classifies samples by counting gates that predict each class.  These networks may be inspected post training to examine the learned logic, however, the authors do not propose a method to interpret the network, which may become arbitrarily large and complex making it transparent in some sense, but un-interpretable under real world conditions.

Our work is aimed at developing a logic based model that uses linear weighting such that the importance of each rule can be understood.  In addition, our approach should be not only transparent, but also interpretable, such that it can be used in practice.  \citeauthor{rudinStopExplainingBlack2019a} clearly articulates three challenges in developing interpretable models \cite{rudinStopExplainingBlack2019a} that align closely to our work. They are described as \textit{logical conditions} (Challenge 1), \textit{linear modeling} (Challenge 2), and \textit{case-based reasoning} (Challenge 3). Logical models, such as decision trees and rule lists, employ logical conditions like ``If-Then'' statements, ``Or'', and ``And'' to make predictions or classify data. These models consist of statements that are combined to form a logical rule, enabling transparent decision-making. Linear modeling, on the other hand, involves assigning scores or weights to features based on their relevance to the model's predictions. This scoring system helps identify the most important features contributing to the model's outcomes. Lastly, case-based reasoning involves explaining the decision-making process for each sample by highlighting the relevant parts of the input data. This approach provides a clear understanding of how the model arrives at its conclusions \cite{rudinStopExplainingBlack2019a}.  Please refer to Table~\ref{tab:RudinResults} to see how the aforementioned methods measure up to these challenges and an assessment of their scalability; our proposed R-NRN address all three challenges.

\section{Neural Reasoning Networks}

In this section, we motivate our \textit{architecture}, Neural Reasoning Network (NRN), and the benefits of our approach to building inherently interpretable AI systems. We then show how NRN is used to instantiate a \textit{supervised classification} algorithm for learning on \textit{tabular data} and close by detailing how this trained network is explained.

Neuro-symbolic AI techniques, such as \cite{DBLP:journals/corr/abs-2006-13155}, have demonstrated capable of producing interpretable models that use logic. LNN showed excellent results for Inductive Logic Programming \cite{Sen_Carvalho_Riegel_Gray_2022}, Entity Linking \cite{jiang-etal-2021-lnn}, KBQA \cite{DBLP:journals/corr/abs-2012-01707}, and reinforcement learning \cite{DBLP:journals/corr/abs-2110-10963}. However, LNN lacks a method for logic induction from data, and cannot leverage GPU acceleration.
We therefore leverage Weighted Lukasiewicz Logic activations functions introduced in \cite{DBLP:journals/corr/abs-2006-13155} to construct a Neural Reasoning Network with ``blocks" of Modified Weighted Lukasiewicz Logic,  which are implemented as PyTorch modules, to produce an interpretable neuro-symbolic network of connected layers that is differentiable, leverages AutoGrad, GPU acceleration, and distributed GPU computation.

\paragraph{NRN} Intuitively, a NRN is similar to a traditional Neural Network (NN), except that each node in the network is interpreted as either an \textit{``And"} (\textit{Conjunction}), or \textit{``Or"} (\textit{Disjunction}) operation as shown in Figure~\ref{Fig:BRNNeuralReasoningNetwork}, which depicts a two-layer Neural Reasoning Network with alternating Conjunction and Disjunction blocks, although different architectures can be easily developed. This interpretation is achieved by representing each node in the network with Weighted Logic that models those operations. Weighted Logic, including Weighted Lukasiewicz Logic introduced in \cite{DBLP:journals/corr/abs-2006-13155}, are a form of fuzzy real valued logic that produce a confidence estimation that the given logical operation is \textit{True}.

\begin{gather}
\label{Equation:ModWeightedConj}
f\bigg(\beta - \sum_j \lvert w_j \rvert \big[1 - [m_jx_j + (1 - m_j)(1 - x_j)]\big]\bigg) \\
\label{Equation:ModWeightedDisj}
f\bigg(1 - \beta + \sum_j \lvert w_j \rvert \big[m_jx_j + (1 - m_j)(1 - x_j)\big]\bigg)
\end{gather}

Formally, a NRN is defined as a composition of a Modified Weighted Lukasiewicz Logic, $f:= g \circ f_k \circ \dots f_2 \circ f_1$, that maps $n$ real valued inputs to $p$ real valued outputs, $f: \mathcal{R}^n \rightarrow \mathcal{R}^p$. Both $\mathcal{R}^n$ and $\mathcal{R}^p$ are in the range $[0, 1]$. Each of the functions $g$ and $f_1 \dots f_k$ can take on a tensor form of one of the two Modified Weighted Lukasiewicz Logic functions corresponding to Conjunction in Equation~\ref{Equation:ModWeightedConj} or Disjunction in Equation~\ref{Equation:ModWeightedDisj} where $\beta$ represents a bias, which is typically fixed and set to 1 to aid interpretation of the network, and for which $\beta \geq 0$. $j$ corresponds to the number of inputs to the logical node, $w \in \mathcal{R}$ is a vector of weights for each input, and $x \in [0, 1]$ is a vector of inputs. Finally, $f$ clamps the result in the range $[0, 1]$ to produce a \textit{truth value}. The key distinction from \cite{DBLP:journals/corr/abs-2006-13155} is $m_j = \mathbf{I}_{w_j > 0}$, which is an indicator identifying if the weight $w_j \in \mathcal{R}$ is greater than 0 and used as a mask in the operation. Negative weights correspond to negation of an input, and the indicator vector acts by flipping those input values to satisfy the negation enabling the network to learn negations without any additional parameters. We discuss details about how this approach enables GPU scaling in the Appendix~\ref{gpu_scaling}.

A ``predicate", $\mathcal{P}$, is the third node type in an NRN and is a mapping between a feature in an input vector $\mathbf{x}$ and a predicate node; the leaves in an NRN.  A single feature may be mapped to multiple predicate nodes.  In addition, predicate nodes contain the feature's description used during explanation of a trained NRN.  Therefore, \textit{one role of predicates is to aid in interpretability of NRN explanations}.

\begin{figure}[t]
    \centering
    \includegraphics[width=1.0\linewidth]{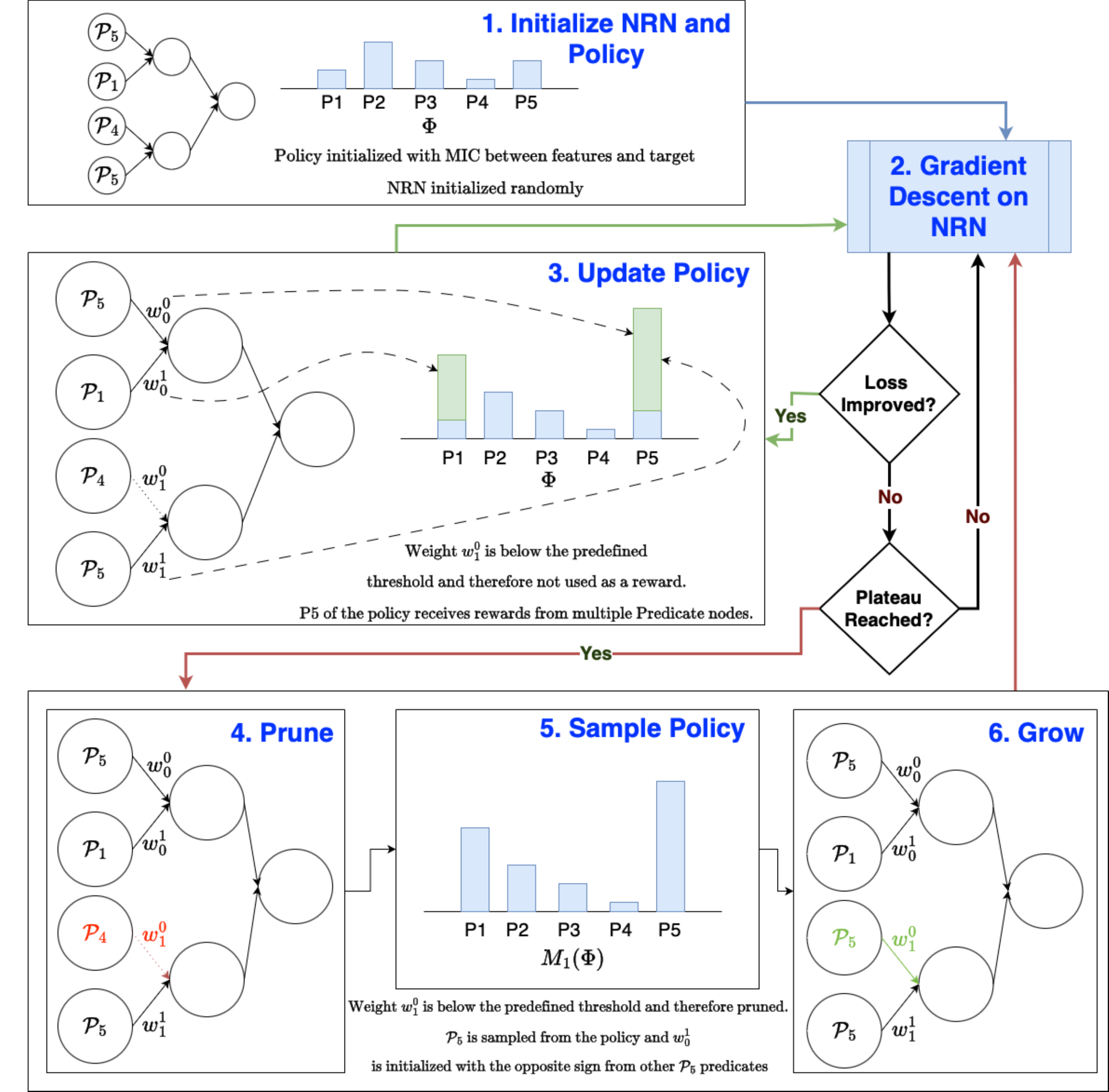}
    \caption{Depiction of iteration using R-NRN logic induction algorithm.}
    \label{fig:BRNAlgoVisual}
\end{figure}

\paragraph{Bandit Reinforced Neural Reasoning Network (R-NRN)} We now proceed to describe our supervised classification algorithm, R-NRN, that induces logic used for prediction from tabular data using NRN. In addition, we will share a description and intuition behind important hyper-parameters and heuristic methods used in the R-NRN algorithm.

R-NRN employs an iterative training algorithm that leverages Gradient Descent and a Multi-Armed Bandit that we depict in Figure~\ref{fig:BRNAlgoVisual} and describe in detail in the Appendix in Algorithm~\ref{Algo:BRN}. Model-1, $M_1(\Phi)$, is a system that generates a proposed NRN, thereby constructing Model-2, $M_2(\Theta)$. Model-2 is subsequently used to predict a target, or multiple targets, and it's parameters, $\Theta$, are updated using Gradient Descent. Finally, if a pre-determined update criteria are met, then a reward function, $R$, computes a reward proportional to the desired performance of Model-2 on the training data and the parameters of Model-1, $\Phi$, are updated with respect to this reward, $r$, using an $\text{UpdatePolicy}$ method for Model-1 . The algorithm is trained for a number of predetermined epochs, $T$, or is stopped using criteria such as reaching a desired performance or a lack of performance improvement.  Next, we describe each step in Figure~\ref{fig:BRNAlgoVisual}.

In \textit{Step 1} of R-NRN training, we initialize Model-1, $M_1(\Phi)$, as a Bayesian UCB Multi-Armed Bandit (BMAB) whose policy is the Maximal Information Coefficient score between features and targets.  The NRN, Model-2, is initialized with a fixed structure resembling that of Figure~\ref{Fig:BRNNeuralReasoningNetwork} with blocks of interpretable nodes that alternate between conjunction and disjunction, the ordering of which is controlled by a hyper-parameter. Much like a traditional NN, one pre-defines the number of logical nodes in each layer, the number of inputs to the logical nodes in each layer, and the number of layers. On initialization, inputs to each logical node are randomly selected from the previous layer.

Following initialization, \textit{Step 2} begins the training procedure and updates weights on each edge in the network using Gradient Descent.  The training proceeds in a typical fashion computing loss over mini-batches until an epoch completes.  

On each iteration the loss is evaluated, and if it's improved, we complete \textit{Step 3} in which the BMAB policy updates with a reward appropriate for generating an improved Model-2 as training continues. We experiment with several reward functions for the BMAB. The \textit{class} strategy, depicted in Figure~\ref{fig:BRNAlgoVisual}, corresponds to the sum of the learned weights for each feature. While fast to compute, this strategy determines rewards at a feature level, rather than a logic level. To enable rewards at a logic level, we developed two additional reward strategies detailed in the Appendix~\ref{RewardStrategies}.

The decision of when to prune the existing logic and generate new logic (\textit{Steps 4, 5, and 6}) is a central part of the R-NRN algorithm and is triggered by plateaus in model performance. As training continues, the learned logic should be closer to optimal and the structure may require less frequent updates. We therefore control the NRN structure update process using three hyper-parameters: $\kappa$, $\tau$, and $\iota$. The hyper-parameter, $\kappa$, sets the number of epochs for which improvement has plateaued that will trigger the model to prune the existing logic. The next two hyper-parameters act together to increase the value of $\kappa$, which is helpful to reduce the frequency of pruning as training progresses. $\iota$ indicates the number of additional epochs to add to $\kappa$, once $\tau$ is reached.

Once the performance plateau is reached, the structure of the NRN is updated with \textit{Steps 4, 5, and 6}. In this update, R-NRN performs a logic pruning and generation process in which predicates with weights above a threshold are kept and those with weights below that threshold are pruned (\textit{Step 4}). New features are sampled from  the BMAB policy, modified by $\delta$ to decrease the likelihood of sampling un-pruned features (\textit{Step 5}).  Finally, those sampled features are placed into the predicate nodes that were pruned (\textit{Step 6}).  We note that multiple predicate nodes may reference the same feature but in separate Conjunction or Disjunction nodes.  Additionally, if the newly generated predicate references a feature already referenced in the un-pruned predicate set, the weights for that predicate are re-initialized with a sign opposite to the average weight across all logics with un-pruned predicates referencing that feature. The effect of flipping the weight is to introduce the negated version of the un-pruned predicate, enabling the Model-2 NRN to represent $l \leq x \leq u$ and $x \leq l  \ \& \ u \leq x $. The details of this procedure are in the Appendix in Algorithm~\ref{Algo:BRNSampling}.

Once trained, the learned structure and weights reveal the induced logic and can be inspected with our explanation algorithm, described next.

\begin{algorithm}[!t]
    \begin{algorithmic}[1]
        \STATE initialize $l = NRN.root$, $t = NRN(x)$, $n = \text{False}$
        \STATE $\textbf{Procedure:}\ expl(l,\ t,\ n):$
            \STATE \hskip1.0em initialize $\epsilon = ``"$
            \STATE \hskip1.0em \textbf{for}$\ j, C \ \textbf{in}\ enumerate(l.children)$
                \STATE \hskip1.0em \hskip1.0em \textbf{if} $n$
                    \STATE \hskip1.0em \hskip1.0em \hskip1.0em $t = 1 - t$
                \STATE \hskip1.0em \hskip1.0em $v_{c} \gets ComputeRequiredChildValue(C, t)$
                \STATE \hskip1.0em \hskip1.0em \textbf{if} $n$
                    \STATE \hskip1.0em \hskip1.0em \hskip1.0em $M = C.value \times \lvert l.weights_j \rvert < v_{c}$
                \STATE \hskip1.0em \hskip1.0em \textbf{else}
                    \STATE \hskip1.0em \hskip1.0em \hskip1.0em $M = C.value \times \lvert l.weights_j \rvert \geq v_{c}$
                \STATE \hskip1.0em \hskip1.0em \textbf{if}$\ M$
                    \STATE \hskip1.0em \hskip1.0em \hskip1.0em \textbf{if}$\ HasChildren(C)$
                        \STATE \hskip1.0em \hskip1.0em \hskip1.0em \hskip1.0em \textbf{if}$\ l.weights_j< 0$
                            \STATE \hskip1.0em \hskip1.0em \hskip1.0em \hskip1.0em \hskip1.0em $n = \textbf{not} \ n$
                        \STATE \hskip1.0em \hskip1.0em \hskip1.0em \hskip1.0em $\epsilon \gets Cat(\epsilon, AddN(expl(C,\ v_{c},\ n), l, n, j))$
                    \STATE \hskip1.0em \hskip1.0em \hskip1.0em \textbf{else}
                        \STATE \hskip1.0em \hskip1.0em \hskip1.0em \hskip1.0em \textbf{if}$\ l.weights_j < 0$
                            \STATE \hskip1.0em \hskip1.0em \hskip1.0em \hskip1.0em \hskip1.0em $v_c = 1 - v_{c}$
                        \STATE \hskip1.0em \hskip1.0em \hskip1.0em \hskip1.0em $\epsilon_{\mathcal{P}} \gets AddN(Cat(C.name, \geq, v_{c}), l, n, j)$
                        \STATE \hskip1.0em \hskip1.0em \hskip1.0em \hskip1.0em $\epsilon \gets Cat(\epsilon, \epsilon_{\mathcal{P}})$
            \STATE \hskip1.0em \textbf{if}$\ \epsilon <> ``"$
                \STATE \hskip1.0em \hskip1.0em $\textbf{return}\ Cat(l.logic\_type, ``(", \epsilon, ``)")$
            \STATE \hskip1.0em $\textbf{return}\ ``"$
    \end{algorithmic}
    \caption{Produce unsimplified explanation}
    \label{Algo:ExplainHighLevel}
  \end{algorithm}

\paragraph{Explanation generation} As mentioned, each node in an NRN is interpretable as an ``And" or ``Or", and we examine a \textit{trained} network to generate an explanation of the model. To interpret the trained network, we implement an algorithm that performs a depth first traversal of an NRNs nodes to produce a sample level explanation. The recursive explanation algorithm consists of evaluating the Weighted Lukasiewicz Logic activation functions of each node to identify the value for each input to a logical node that is required for that input to produce the nodes \textit{truth value} given all other input values, as in Algorithm~\ref{Algo:ExplainHighLevel}, line 7; and surfacing only those nodes for which the specific sample value meets this threshold, lines 8-11.  In Algorithm~\ref{Algo:ExplainHighLevel}, $l$ is the root node of an NRN, $t$ is the output from a trained NRN, and $n$ is a boolean assisting with handling negations during traversal.  Appendix~\ref{ExplanationGeneration} includes details of procedures for $ComputeRequiredChildValue$, $AddN$ which adds ``NOT" to the explanation string if required, $Cat$ which performs string concatenation, and $expl$, which produces a string form of the explanation.

\subsubsection{Explanation simplification} The logic extracted with Algorithm~\ref{Algo:ExplainHighLevel} may be complex and we therefore perform a simplification process, which is of critical importance to ensure NRNs are interpretable in practice. The simplification process implies applying logical rules to shorten and simplify the explanations. Our goal is to make the explanations as simple as possible while preserving all the logic of the model. 
As in Algorithm~\ref{Algo:ExplainationSimplification}, we apply several recursive sub-algorithms until the following conditions are met 1) standardize the logical structure to conjunctive (or disjunctive) normal form that only contains negations at the leaves -- represented by ``predicates" (line 2) by traversing the explanation, represented as a tree, and applying logical rules to push negations down to the leaves level, 
2) identify propositional equivalences and apply carefully crafted simplification rules to remove unnecessary logic such as collapsing sequences of consecutive conjunctions or disjunctions to a single operation, and removing redundant predicates (lines 3, 4, 5, 6), and 3) aggregate all the conditions which are true for the sample into a single conjunction while ignoring all the logic parts which do not hold for the sample (lines 7, 8). The resulting sample explanation is a single conjunction of rules as shown in Figure~\ref{fig:ExplanationExample}.  
Additional details of the simplification functions and rules are located in the Appendix section~\ref{ExplanationSimplification} and additional example outputs are shown in section~\ref{ExplanationExamples}. 

\begin{algorithm}[t]
  \begin{algorithmic}[1]
\STATE initialize $n = Explanation.root$
\\
\STATE $push\_negations\_down(n)$ \COMMENT{(Rules \# 1, \# 4, \# 5, \# 6)}
\STATE $collapse\_repeated\_operands(n)$ \COMMENT{(Rules \# 7, \# 8)}
\STATE $remove\_redundant\_predicates(n)$ \COMMENT{(Rules \# 2, \# 3)}
\STATE $collapse\_single\_operands(n)$ \COMMENT {(Rule \# 9)}
\STATE $remove\_redundant\_predicates(n)$ \COMMENT{(Rules \# 2, \# 3)}
\STATE $collapse\_sample\_explanation(n)$
\STATE $remove\_redundant\_predicates(n)$ \COMMENT{(Rules \# 2, \# 3)}
\end{algorithmic}
\caption{Explanation simplification}
\label{Algo:ExplainationSimplification}
\end{algorithm}

\begin{figure}[!t]
    \centering
    \includegraphics[width=1.0\linewidth]{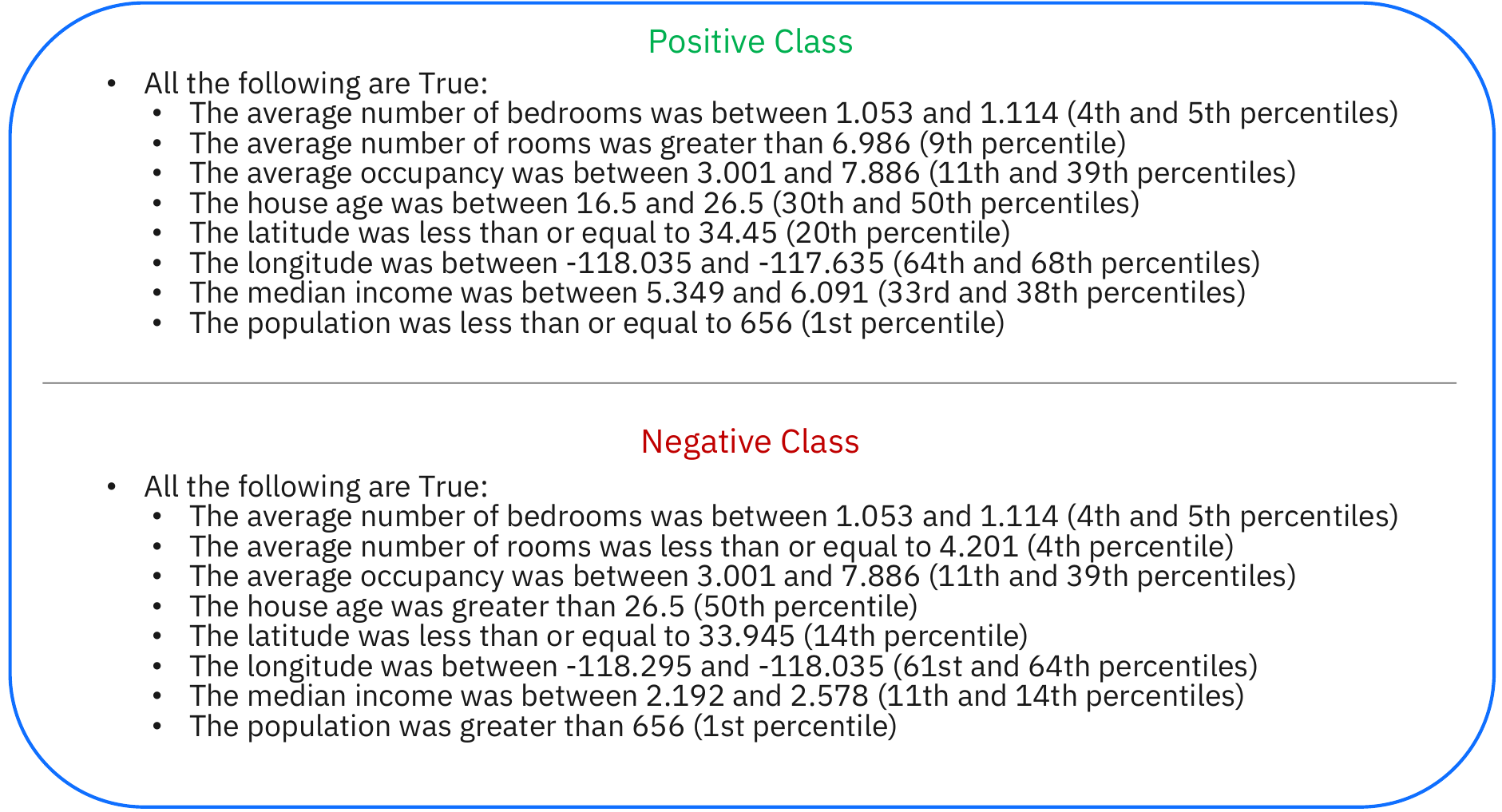}
    \caption{Example sample explanations from trained R-NRN on the \texttt{california} dataset.  Shows 100\% of the simplified model explanation.}
    \label{fig:ExplanationExample}
\end{figure}

\begin{table*}[!t]
\centering
    \begin{tabular}{p{19mm}p{11mm}|p{9.5mm}p{10mm}p{9.5mm}p{9.5mm}p{10mm}p{9.5mm}p{10mm}p{9.5mm}p{9.5mm}p{10.5mm}}
    \toprule
     & R-NRN        & BRCG                    & DIF                  & NAM                               & CFN                               & FTT                               & DAN                     & MLP                               & RF                                 & XGB                                  & GBT \\ \midrule
    Average & 0.980           & 0.913\tsp{\up  7\%}  & 0.844\tsp{\up  16\%} & \underline{1.028}\tsp{\down  5\%} & 0.991\tsp{\down  1\%}             & 0.763\tsp{\up  28\%}              & 0.959\tsp{\up  2\%}   & 0.865\tsp{\up  13\%}              & 1.000\tsp{\down 2\%}               & 0.979\tsp{\up  0\%}                  & 0.993\tsp{\down  1\%} \\ 
    Avg. Time    & 114          & 143          & 77           & 189             & 213          & 293              & 176             & 174             & 37     & \textbf{10} & 21  \\
    Avg. Params    & \textbf{1}          & -          & 6           & 861             & 37         & 182              & 1463             & 62             & -     & - & -  \\
    \multicolumn{2}{l|}{Ours vs W/L/T}            & 18/0/4                & 21/0/1               & 0/20/2                            & 4/9/9                             & 20/2/0                            & 8/2/12                  & 12/2/8                            & 5/12/5                             & 5/11/6                               & 7/9/6  \\
    \multicolumn{2}{l|}{Ours vs p-value}         & $0.047^{*}$                & $0.001^{**}$                & 0.130                             & 0.614                             & $0.000^{***}$                            & 0.460                   & $0.027^{*}$                             & 0.734                              & 0.769                                & 0.991  \\
    \bottomrule
    \end{tabular}
\caption{Experimental results showing average test AUC of 5 models trained with different random seeds and \textit{normalized to RF scores}. \underline{Underline} is highest average test AUC. Average trial time in seconds and average number of parameters in thousands (for neural methods, otherwise ``-") computed over a benchmark of 22 tabular classification datasets \cite{grinsztajn2022why} across a variety of algorithms. We show Win/Loss/Ties pairwise comparisons and p-values from two-tailed Mann-Whitney U test comparing R-NRN to other algorithms. Average test score of R-NRN better \tsp{\up  \%} / worse \tsp{\down  \%} vs Other.}
\label{Table:ExperimentalResults}
\end{table*}

\section{Results}

In this section we present the results of an extensive empirical study comparing R-NRN to tree-based algorithms, traditional deep learning, and specialized deep learning models designed for tabular data. In addition, we present example explanations from R-NRN and evaluate their quality.

\paragraph{Predictive performance} To evaluate our method, we train R-NRN, as well as one or more representative models from each class of models described in the related work, on the benchmark for tabular datasets proposed by \cite{grinsztajn2022why} detailed in Appendix~\ref{Datasets}. The authors performed extensive analysis on these datasets to ensure they represent a variety of problems that are not easily solved and are similar to a typical application of AI/ML to real world tabular data. We conduct our own hyper-parameter search with baseline models using the same hyper-parameter ranges as in \cite{grinsztajn2022why} and a larger validation split size, as the method described in \cite{grinsztajn2022why} used too small a validation set for optimal hyper-parameter selection and did not use the methods described in \cite{DBLP:journals/corr/abs-2106-11189} to optimally tune the MLP based models.

Our study focuses on classification and we evaluate all algorithms using micro averaged AUC so as to require good performance on all classes and to not impact performance with less than optimal decision boundary selection. The results of our hyper-parameter search, shown in Table~\ref{Table:ExperimentalResults}, show the average test AUC score \textit{normalized to the AUC for Random Forest} for each algorithm from 5 models trained with different random seeds (1 to 5) using the best hyper-parameters identified with an extensive, approximately 400 trial search guided by a TPESampler \cite{pmlr-v28-bergstra13} from the Optuna \cite{optuna_2019} library. For R-NRN and MLP, we allow the TPESampler to choose if FeatureBinarizationFromTrees \cite{aix360-sept-2019} pre-processing is used, explained further in Appendix~\ref{HyperParamAnalysis}. For Boolean Rules via Column Generation (BRCG), DiffLogic (DIF), NAM, CoFrNet (CFN), DANet (DAN), and FT-Transformer (FTT) we choose hyper-parameter ranges suggested by the authors if available, or infer the ranges from the author's code bases.

We examine the average scores\footnote{DIF reported using only seed 42. All tuning used seed 42 and DIF test AUC is random for all other seeds. \label{foot:DiffLogicAUC}} along with the results of the two-tailed Mann-Whitney U test with $\alpha = 0.05$ and find that R-NRN performs similarly to XGB, RF, GBT, NAM, CFN, and DAN. R-NRN also outperforms other logic based methods BRCG (+7\%) and DIF (+16\%), some neural methods designed for tabular data FTT (+28\%), and traditional deep learning MLP (+13\%). 

\paragraph{Explanation quality}

Quantitative evaluations of model explanation are aimed at producing measures that indicate the quality of a produced explanation and can be defined in a variety ways, each measuring different aspects of an explanation \cite{DBLP:journals/corr/abs-2201-08164}.

Size is a measure of explanation compactness and gives information regarding the presentation of the explanation \cite{DBLP:journals/corr/abs-2201-08164}.  Smaller explanations are generally considered better since those will be easiest to understand, although computing the size of an explanation is dependent on the explanations format.  For NAM, RF, and SHAP we count features, since an explanation consists of a feature importance value for each input feature.  For R-NRN, we compute explanation size as the average of the count of rules produced by a sample of explanations on the test set.

Our second metric, Single Deletion, is a measure of correctness, which 
is an evaluation of how faithfully the explanation describes the model \cite{DBLP:journals/corr/abs-2201-08164}.  A Single Deletion score can be calculated for any algorithm that produces feature importance; for R-NRN see Appendix~\ref{FeatureImportanceComp}.  We report both Spearman and Pearson Correlation between the feature importance score produced by each algorithm and the change in AUC on the test set when that single feature is perturbed.  

While there are many more metrics, these two provide essential information when evaluating explanation quality.  Namely, can a human reasonably understand the explanation (size), and can a human trust that the explanation is correctly describing the model (single deletion).  Table~\ref{Table:ExplanationQuality} shows R-NRN explanations are 31\% smaller, while also more accurate in terms of Single Deletion when compared with both NAM and RF.  While Single Deletion Pearson Correlation is better for SHAP applied to RF, R-NRN explanations are generated 96\% faster than SHAP on RF.  We choose these algorithms as our baselines since the comparison covers other inherently interpretable methods (NAM), our best performing deep-tree-ensemble model (RF), and a post-hoc explanation technique (SHAP).  

Figure~\ref{fig:ExplanationExample} presents an example of sample level explanation produced by R-NRN on the California dataset for qualitative analysis.  This data consists of predicting if the price of a house is above or below the median and includes features such as the number of rooms, number of bedrooms, latitude, longitude etc.  The example compares randomly selected explanations for the positive and negative class and demonstrates the ease with which one can interpret explanations using R-NRN. Furthermore, the explanations match our intuition about the factors that impact housing prices.

\section{Conclusion}

\paragraph{Limitations} R-NRN shows good performance on the benchmark datasets for tabular classification, but it is not yet clear if this approach will work for regression or other fundamental problems in AI such as with more complex domains like Computer Vision or Natural Language Processing.  Furthermore, the flexibility to modify the initialized structure is somewhat limited in R-NRN, raising questions as to if this approach can be used on such complex domains.  Also, while we present examples of explanations produced by R-NRN as well as quantitative analysis of their quality, we do not perform a user-study that would give more insight into the usefulness of such explanations in practice.

\begin{table}
\centering
    
    \begin{tabular}{p{19.5mm}p{11mm}|p{9.5mm}p{9.5mm}p{13.0mm}}
    \toprule
     & R-NRN & NAM & RF & SHAP(RF) \\ \midrule
     Size & 14.8 & 21.3\tsp{\downbetter 31\%} & 21.3\tsp{\downbetter 31\%} & 21.3\tsp{\downbetter 31\%} \\
     SD Spearman & 0.80 & 0.74\tsp{\up 8\%} & 0.73\tsp{\up 10\%} & 0.79\tsp{\up 1\%} \\
     SD Pearson & 0.83 & 0.74\tsp{\up 12\%} & 0.80\tsp{\up 4\%} & 0.91\tsp{\down 9\%} \\
     Avg. Time & 270 & 194\tsp{\upworse 39\%} & 0.0\tsp{\upworse N/A} & 6005\tsp{\downbetter 96\%} \\
    \bottomrule
    \end{tabular}
\caption{Explanation evaluation with \textit{Size} and \textit{Single Deletion} Spearman/Pearson Correlation, and average time in seconds to compute feature importance, reported against benchmark datasets \cite{grinsztajn2022why}.  R-NRN \textit{Size} and \textit{Avg. Time} better \tsp{\downbetter  \%} / worse \tsp{\upworse  \%} vs Other.  R-NRN \textit{Single Deletion} score better \tsp{\up  \%} / worse \tsp{\down  \%} vs Other.}
\label{Table:ExplanationQuality}
\end{table}

\paragraph{Conclusion} NRNs are used to construct a supervised classification algorithm, R-NRN, with performance on tabular classification similar to RF, XGB, and GBT. R-NRN explanations are smaller and more accurate compared to feature importance based approaches including RF and NAM. Future work could leverage NRNs to develop inherently interpretable algorithms for NLP, computer vision, and reasoning domains.

\section*{Acknowledgement}
A special thanks to Naoki Abe for sharing his experience and knowledge with us along our way.

\bibliography{bibliography}

\appendix

\section{Hyper-parameter tuning}
\label{Hyper-parameter tuning}

\subsubsection{Overview}For each algorithm, we perform approximately 400 trials hyper-parameter search using a TPESampler \cite{pmlr-v28-bergstra13} from the Optuna \cite{optuna_2019} package. All tuning is performed with random seed set to 42. The hyper-parameter ranges for each search are listed in Table~\ref{Table2}, Table~\ref{Table3}, Table~\ref{Table4}, Table~\ref{Table5}, Table~\ref{Table7}, Table~\ref{Table9}, Table~\ref{Table6}, Table~\ref{Table8}, Table~\ref{Table10}, Table~\ref{Table11}. 

\subsubsection{Train/val/test splits}
Following a procedure based on \cite{grinsztajn2022why}, we use 60\% of the samples as our Train split and limit the training set size to 10,000 samples using random sampling. We split the remaining samples including 50\% in our Validation split and 50\% in our Test split. Validation and Test data are truncated using random sampling to 50,000 samples for speed of experiments.

\subsubsection{Cross validation} We use the cross validation strategy for hyper-parameter optimization from \cite{grinsztajn2022why} as follows:

\begin{itemize}
    \item If we have more than 6000 samples we use 1 fold.
    \item If there are 3000 to 6000 samples we use 2 folds.
    \item If there are 1000 to 3000 samples we use 3 folds.
    \item If there are less than 1000 samples we use 5 folds.
\end{itemize}

\subsubsection{Comparison to other experiments} Our results show that several neural models are competitive with tree-based methods and thus differ from other recent studies on neural networks for tabular data such as \cite{9998482}, \cite{cherepanova2023a}, \cite{grinsztajn2022why}, \cite{gorishniy2021revisiting}. This may due in part to our use of AUC rather than accuracy. AUC performance is not dependent on the selections of an appropriate decision boundary in classification problems and thus removes any performance gain/loss from this hyper-parameter. Furthermore, those studies did not include, or did not perform full tuning and experimentation with the neural methods that we discover to be competitive with tree-based algorithms.

Our results for RF, XGB and GBT are somewhat different compared to \cite{grinsztajn2022why}.  This could be due to the use of AUC rather than accuracy, as well as using a larger validation set and performing hyper-parameter tuning with TPESampling rather than random search.

\section{Compute Hardware}

Tree-based models were trained on CPU. R-NRN, DiffLogic, NAM, DANet, CoFrNet, MLP, and FTT, were trained on GPU (V100) with the PyTorch \cite{NEURIPS2019_9015} CosineAnnealingWarmRestarts scheduler \cite{DBLP:journals/corr/LoshchilovH16a}. FTT required an A100 GPU for datasets 44156 to 44162. For all algorithms we use 3 CPU and 48GB of RAM. For MLP, DANet, CoFrNet, DiffLogic, and R-NRN we fix the maximum number of epochs at 200. For FTT we are unable to perform 200 epochs due to the size and slow speed of the model and therefore fix the maximum epochs to 50. NAM's authors reported using 1000 epochs and we do the same. The ranges for hyper-parameter search are available in the Appendix and closely follow the ranges used by \cite{grinsztajn2022why}.

\section{Algorithm Implementations}  We use the following implementations for each algorithm.  These are either the official implementations as provided by the original authors or re-implementations from sources like scikit-learn and github repos with significant usage as demonstrated by stars.

\begin{links}
    \link{FT-Transformer}{https://github.com/lucidrains/tab-transformer-pytorch}
    \link{Neural Additive Models}{https://github.com/google-research/google-research/tree/master/neural_additive_models}
    \link{CoFrNet}{https://github.com/Trusted-AI/AIX360}
    \link{DANet}{https://github.com/WhatAShot/DANet}
    \link{BRCG}{https://github.com/Trusted-AI/AIX360}
    \link{Random Forest}{https://scikit-learn.org/stable/index.html}
    \link{Gradient Boosted Trees}{https://scikit-learn.org/stable/index.html}
    \link{XGBoost}{https://xgboost.readthedocs.io/en/stable/}
\end{links}

MLP is implemented within our open-sourced code

\section{Datasets}
\label{Datasets}

Table \ref{tbl:Bench-Datasets} lists the classification datasets obtained from \cite{grinsztajn2022why}.  While the benchmark provided by the original authors is more extensive, also including regression datasets, our work is focused only on classification and therefore we list only those datasets.

\begin{table*}[ht]
  \small
  \centering
  \begin{tabular}{@{}llrrl@{}}
    \toprule \textbf{OpenML ID} & \textbf{Name} & \textbf{n\_samples} & \textbf{n\_features} & \textbf{Link} \\ \midrule
44089 & credit & 16714 & 10 & \url{https://www.openml.org/d/44089} \\
44090 & california & 20634 & 8 & \url{https://www.openml.org/d/44090} \\
44091 & wine & 2554 & 11 & \url{https://www.openml.org/d/44091} \\
44120 & electricity & 38474 & 7 & \url{https://www.openml.org/d/44120} \\
44121 & covertype & 566602 & 10 & \url{https://www.openml.org/d/44121} \\
44122 & pol & 10082 & 26 & \url{https://www.openml.org/d/44122} \\
44123 & house\_16H & 13488 & 16 & \url{https://www.openml.org/d/44123} \\
44124 & kdd\_ipums\_la\_97-small & 5188 & 20 & \url{https://www.openml.org/d/44124} \\
44125 & MagicTelescope & 13376 & 10 & \url{https://www.openml.org/d/44125} \\
44126 & bank-marketing & 10578 & 7 & \url{https://www.openml.org/d/44126} \\
44127 & phoneme & 3172 & 5 & \url{https://www.openml.org/d/44127} \\
44128 & MiniBooNE & 72998 & 50 & \url{https://www.openml.org/d/44128} \\
44129 & Higgs & 940160 & 24 & \url{https://www.openml.org/d/44129} \\
44130 & eye\_movements & 7608 & 20 & \url{https://www.openml.org/d/44130} \\
44131 & jannis & 57580 & 54 & \url{https://www.openml.org/d/44131} \\
44156 & electricity & 38474 & 8 & \url{https://www.openml.org/d/44156} \\
44157 & eye\_movements & 7608 & 23 & \url{https://www.openml.org/d/44157} \\
44158 & KDDCup09\_upselling & 5032 & 45 & \url{https://www.openml.org/d/44158} \\
44159 & covertype & 423680 & 54 & \url{https://www.openml.org/d/44159} \\
44160 & rl & 4970 & 12 & \url{https://www.openml.org/d/44160} \\
44161 & road-safety & 111762 & 32 & \url{https://www.openml.org/d/44161} \\
44162 & compass & 16644 & 17 & \url{https://www.openml.org/d/44162} \\ \bottomrule
  \end{tabular}
  \caption{Classification benchmark datasets}
  \label{tbl:Bench-Datasets}
\end{table*}

\section{Dataset Level Results}

Table~\ref{Table:ExperimentalResultsDetailed} shows the result for our experiments on 22 open-source dataset at the data set level.  We include the performance of the algorithms on each dataset relative to our approach, R-NRN, as well as the results of a pairwise comparison of each algorithm using a one-sided Mann-Whitney U test for each dataset.  We use this approach to determine if any single algorithm has an average test score that is greater than all other algorithms.

\begin{table*}
\centering
    \caption{Experimental results showing average test AUC of 5 models trained with different random seeds and \textit{normalized to RF scores}. \underline{Underline} is highest average test AUC. \textbf{Bold} is statistically significantly better than all other algorithms at $\alpha = 0.05$ using one sided Mann-Whitney U test. R-NRN better \tsp{\up  \%} / worse \tsp{\down  \%} vs Other.}
    \begin{tabular}{p{10mm}p{12mm}|p{10.5mm}p{10.5mm}p{10.5mm}p{10.5mm}p{10.7mm}p{10.5mm}p{10.5mm}p{10.5mm}p{10.5mm}p{10.5mm}}
    \toprule
    OpenML ID & R-NRN        & BRCG                    & DIF                  & NAM                               & CFN                               & FTT                               & DAN                     & MLP                               & RF                                 & XGB                                  & GBT \\ \midrule
    44089 & \underline{1.060} & 0.980\tsp{\up  8\%}  & 0.822\tsp{\up  29\%} & 1.057\tsp{\up  0\%}               & 1.046\tsp{\up  1\%}               & 0.642\tsp{\up  65\%}              & 0.995\tsp{\up  7\%}   & 0.939\tsp{\up  13\%}              & 1.000\tsp{\up  6\%}                & 0.679\tsp{\up  56\%}                 & 1.001\tsp{\up  6\%} \\
    44090 & 0.990             & 0.899\tsp{\up 10\%}  & 0.891\tsp{\up  11\%} & 1.065\tsp{\down  7\%}             & 1.024\tsp{\down  3\%}             & 0.937\tsp{\up  6\%}               & 0.976\tsp{\up  1\%}   & \underline{1.066}\tsp{\down  7\%} & 1.000\tsp{\down 1\%}               & 1.012\tsp{\down  2\%}                & 1.010\tsp{\down  2\%} \\
    44091 & 0.974             & 0.905\tsp{\up  8\%}  & 0.912\tsp{\up  7\%}  & \underline{1.017}\tsp{\down  4\%} & 1.005\tsp{\down  3\%}             & 0.707\tsp{\up  38\%}              & 0.940\tsp{\up  4\%}   & 0.963\tsp{\up  1\%}               & 1.000\tsp{\down 3\%}               & 0.997\tsp{\down  2\%}                & 0.983\tsp{\down  1\%} \\
    44120 & 0.944             & 0.885\tsp{\up  7\%}  & 0.802\tsp{\up  18\%} & \textbf{1.022}\tsp{\down  8\%}    & 0.969\tsp{\down  3\%}             & 0.440\tsp{\up  115\%}             & 0.935\tsp{\up  1\%}   & 0.653\tsp{\up  45\%}              & 1.000\tsp{\down 6\%}               & 1.002\tsp{\down  6\%}                & 1.013\tsp{\down  7\%} \\
    44121 & 0.959             & 0.885\tsp{\up  8\%}  & 0.722\tsp{\up  33\%} & 0.985\tsp{\down  3\%}             & 0.983\tsp{\down  2\%}             & 0.699\tsp{\up  37\%}              & 0.950\tsp{\up  1\%}   & 0.778\tsp{\up  23\%}              & \underline{1.000}\tsp{\down 4\%}   & 0.999\tsp{\down  4\%}                & 0.951\tsp{\up  1\%} \\
    44122 & 0.980             & 0.976\tsp{\up  0\%}  & 0.851\tsp{\up  15\%} & 1.007\tsp{\down  3\%}             & \underline{1.014}\tsp{\down  3\%} & 0.595\tsp{\up  65\%}              & 0.961\tsp{\up  2\%}   & 1.003\tsp{\down  2\%}             & 1.000\tsp{\down 2\%}               & 1.003\tsp{\down  2\%}                & 1.003\tsp{\down  2\%} \\
    44123 & 1.041             & 0.938\tsp{\up  11\%} & 0.837\tsp{\up  24\%} & \textbf{1.069}\tsp{\down  3\%}    & 1.039\tsp{\up  0\%}             & 0.790\tsp{\up  32\%}              & 0.996\tsp{\up  4\%}   & 1.030\tsp{\up  1\%}               & 1.000\tsp{\up 4\%}                 & 1.007\tsp{\up  3\%}                  & 1.004\tsp{\up  4\%} \\
    44124 & 1.062             & 1.001\tsp{\up  6\%}  & 0.984\tsp{\up  8\%}  & \underline{1.067}\tsp{\down  0\%} & 1.015\tsp{\up  5\%}               & 1.011\tsp{\up  5\%}               & 0.996\tsp{\up  7\%}   & 1.052\tsp{\up  1\%}               & 1.000\tsp{\up 6\%}                 & 1.001\tsp{\up  6\%}                  & 1.002\tsp{\up  6\%} \\
    44125 & 1.004             & 0.912\tsp{\up  10\%} & 0.930\tsp{\up  8\%}  & \textbf{1.052}\tsp{\down  5\%}    & 1.039\tsp{\down  3\%}             & 0.590\tsp{\up  70\%}              & 0.995\tsp{\up  1\%}   & 0.634\tsp{\up  58\%}              & 1.000\tsp{\up 0\%}                 & 1.001\tsp{\up  0\%}                  & 0.999\tsp{\up  0\%} \\
    44126 & 0.995             & 0.965\tsp{\up  3\%}  & 0.930\tsp{\up  7\%}  & \textbf{1.096}\tsp{\down  9\%}    & 1.051\tsp{\down  5\%}             & 0.800\tsp{\up  24\%}              & 0.988\tsp{\up  1\%}   & 0.993\tsp{\up  0\%}               & 1.000\tsp{\down 1\%}               & 1.007\tsp{\down  1\%}                & 1.003\tsp{\down  1\%} \\
    44127 & 0.948             & 0.925\tsp{\up  3\%}  & 0.802\tsp{\up  18\%} & 1.005\tsp{\down  6\%}             & 0.929\tsp{\up  2\%}               & \underline{1.013}\tsp{\down  6\%} & 0.969\tsp{\down  2\%} & 0.966\tsp{\down  2\%}             & 1.000\tsp{\down 5\%}               & 0.999\tsp{\down  5\%}                & 0.980\tsp{\down  3\%} \\
    44128 & 1.011             & 0.945\tsp{\up  7\%}  & 0.921\tsp{\up  10\%} & \textbf{1.027}\tsp{\down  2\%}    & 0.961\tsp{\up  5\%}               & 0.938\tsp{\up  8\%}               & 1.009\tsp{\up  0\%}   & 0.671\tsp{\up  51\%}              & 1.000\tsp{\up 1\%}                 & 1.011\tsp{\up  0\%}                  & 1.006\tsp{\up  0\%} \\
    44129 & 0.982             & 0.886\tsp{\up  11\%} & 0.760\tsp{\up  29\%} & \textbf{1.075}\tsp{\down  9\%}    & 0.961\tsp{\up  2\%}               & 0.868\tsp{\up  13\%}              & 0.964\tsp{\up  2\%}   & 0.713\tsp{\up  38\%}              & 1.000\tsp{\down 2\%}               & 1.007\tsp{\down  3\%}                & 0.999\tsp{\down  2\%} \\
    44130 & 0.922             & 0.872\tsp{\up  6\%}  & 0.847\tsp{\up  9\%}  & 0.947\tsp{\down  3\%}             & 0.895\tsp{\up  3\%}               & 0.860\tsp{\up  7\%}               & 0.881\tsp{\up  5\%}   & 0.926\tsp{\down  0\%}               & 1.000\tsp{\down 8\%}               & \underline{1.019}\tsp{\down  10\%}   & 1.016\tsp{\down  9\%} \\
    44131 & 1.013             & 0.914\tsp{\up  11\%} & 0.886\tsp{\up  14\%} & \textbf{1.062}\tsp{\down  5\%}    & 1.042\tsp{\down  3\%}             & 0.945\tsp{\up  7\%}               & 0.965\tsp{\up  5\%}   & 0.997\tsp{\up  2\%}               & 1.000\tsp{\up 1\%}                 & 1.001\tsp{\up  1\%}                  & 0.992\tsp{\up  2\%} \\ 
    44156 & 0.943             & 0.859\tsp{\up  10\%} & 0.736\tsp{\up  28\%} & 1.004\tsp{\down  6\%}             & 0.952\tsp{\down  1\%}             & 0.564\tsp{\up  67\%}              & 0.934\tsp{\up  1\%}   & 0.941\tsp{\up  0\%}               & 1.000\tsp{\down 6\%}               & 1.011\tsp{\down  7\%}                & \underline{1.011}\tsp{\down  7\%} \\
    44157 & 0.883             & 0.857\tsp{\up  3\%}  & 0.854\tsp{\up  3\%}  & 0.941\tsp{\down  6\%}             & 0.905\tsp{\down  2\%}             & 0.732\tsp{\up  21\%}              & 0.873\tsp{\up  1\%}   & 0.834\tsp{\up  6\%}               & 1.000\tsp{\down 12\%}              & \textbf{1.031}\tsp{\down  14\%}      & 0.989\tsp{\down  11\%} \\
    44158 & 1.105             & 1.010\tsp{\up  9\%}  & 0.840\tsp{\up  32\%} & \textbf{1.139}\tsp{\down  3\%}    & 1.115\tsp{\down  1\%}             & 0.650\tsp{\up  70\%}              & 0.972\tsp{\up  14\%}  & 0.904\tsp{\up  22\%}              & 1.000\tsp{\up 11\%}                & 1.014\tsp{\up  9\%}                  & 1.012\tsp{\up  9\%} \\
    44159 & 0.943             & 0.871\tsp{\up  8\%}  & 0.778\tsp{\up  21\%} & 0.992\tsp{\down  5\%}             & 0.992\tsp{\down  5\%}             & 0.714\tsp{\up  32\%}              & 0.983\tsp{\down  4\%} & 0.576\tsp{\up  64\%}              & 1.000\tsp{\down 6\%}               & \textbf{1.006}\tsp{\down  6\%}       & 0.958\tsp{\down  2\%} \\
    44160 & 0.848             & 0.776\tsp{\up  9\%}  & 0.752\tsp{\up  13\%} & 0.956\tsp{\down  11\%}            & 0.873\tsp{\down  3\%}             & 0.897\tsp{\down  5\%}             & 0.863\tsp{\down  2\%} & 0.653\tsp{\up  30\%}              & \underline{1.000}\tsp{\down 15\%}  & 0.999\tsp{\down  15\%}               & 0.984\tsp{\down  14\%} \\
    44161 & 0.978             & 0.948\tsp{\up  3\%}  & 0.927\tsp{\up  5\%}  & \underline{1.054}\tsp{\down  7\%} & 1.047\tsp{\down  7\%}             & 0.660\tsp{\up  48\%}              & 0.992\tsp{\down  1\%} & 0.937\tsp{\up  4\%}               & 1.000\tsp{\down 2\%}               & 1.005\tsp{\down  3\%}                & 0.988\tsp{\down  1\%} \\
    44162 & 0.969             & 0.874\tsp{\up  11\%} & 0.785\tsp{\up  23\%} & 0.983\tsp{\down  1\%}             & 0.951\tsp{\up  2\%}               & 0.736\tsp{\up  32\%}              & 0.958\tsp{\up  1\%}   & 0.808\tsp{\up  20\%}              & \textbf{1.000}\tsp{\down 3\%}      & 0.718\tsp{\up  35\%}                 & 0.945\tsp{\up  3\%} \\ 
    \bottomrule
    \end{tabular}
\label{Table:ExperimentalResultsDetailed}
\end{table*}

\section{Explanation Examples}

\begin{figure*}
    \centering
    \includegraphics[width=1.0\textwidth]{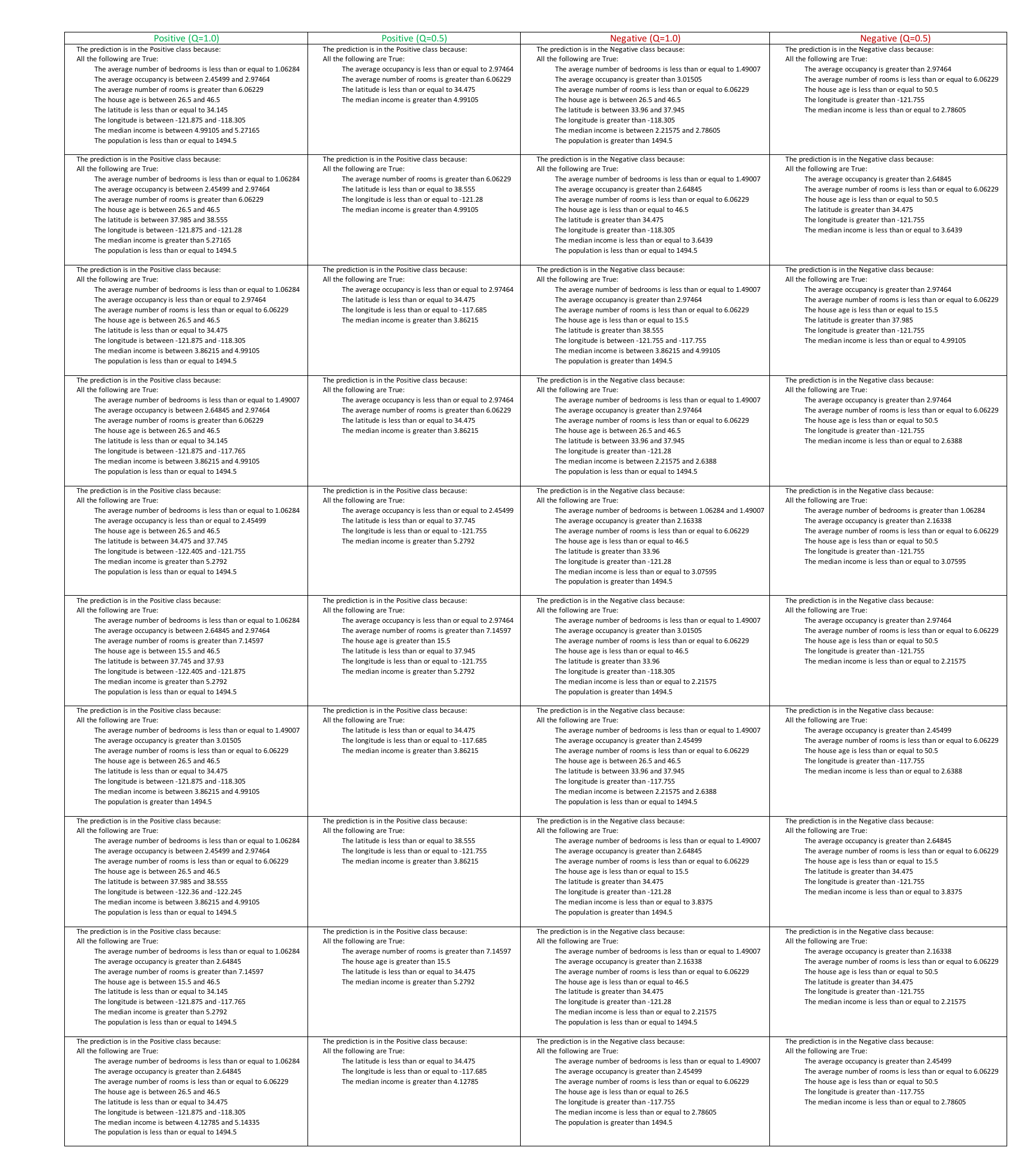}
    \caption{10 randomly selected pairs of positive and negative class explanations produced using R-NRN trained on the \textit{California} dataset.}
    \label{fig:CaliforniaExplanationExamples}
\end{figure*}

Figure~\ref{fig:CaliforniaExplanationExamples} includes 10 randomly sampled positive and negative class sample level explanations for predictions made on the \textit{California} dataset. The pairing of positive and negative class samples bears no significance and is presented in this format purely for ease of contrasting examples from the two classes. Explanations are shown at quantile thresholds of 1.0 and 0.5 corresponding to the entire model and the sub-graph of the Neural Reasoning Network with weights in the top 50\%. Explanation can vary slightly at different quantile thresholds due to sub-graphs of the Neural Reasoning Network falling below the threshold and thus excluded from the explanation simplification process.

\begin{figure*}
    \centering
    \includegraphics[width=0.85\textwidth]{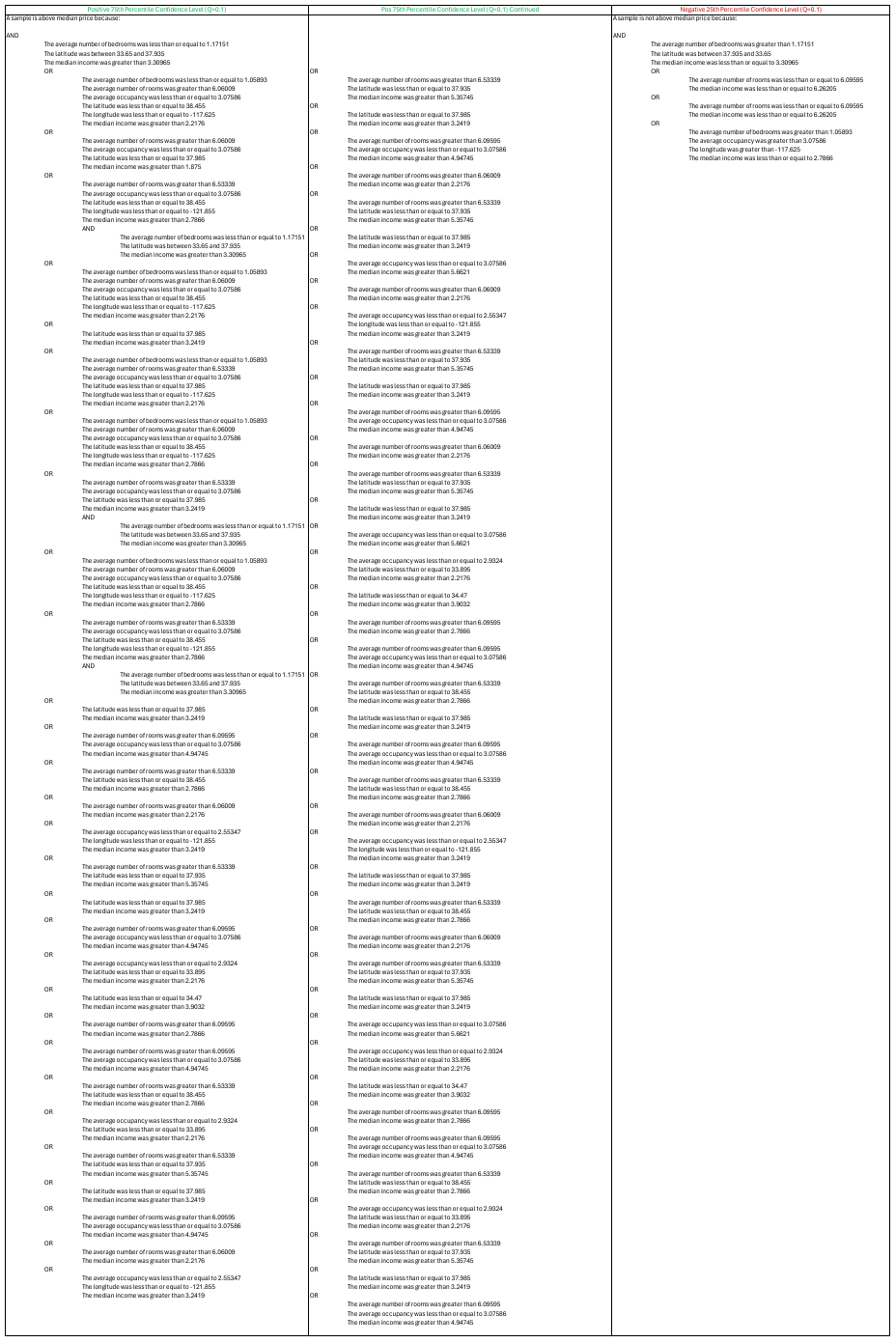}
    \caption{Example of global explanations for the positive and negative class produced using R-NRN trained on the \textit{California} dataset.  We set the confidence level for the positive class explanation to the 75th percentile of test set predictions and the negative class to the 25th percentile.  We set the quantile weight threshold to 0.1, meaning we select logic that falls into the top 10\% in terms of importance.}
    \label{fig:GlobalCaliforniaExplanationExamples}
\end{figure*}

Figure~\ref{fig:GlobalCaliforniaExplanationExamples} shows examples of global explanations for the positive and negative class produced using R-NRN trained on the \textit{California} dataset.  We set the confidence level for the positive class explanation to the 75th percentile of test set predictions and the negative class to the 25th percentile.  We set the quantile weight threshold to 0.1, meaning we select logic that falls into the top 10\% in terms of importance.  This figure demonstrates two important aspects of NRN.  First, we can model complex functions as demonstrated by the size of global explanation of the positive class.  Second, sample explanations can be considerable simplified for ease of interpretation as demonstrated by the comparison of the size of explanations at the global level and sample level ~\ref{fig:CaliforniaExplanationExamples}.

\label{ExplanationExamples}

\section{Explanation Generation} The sample level explanation algorithm is shown in Algorithm~\ref{Algo:ExplainHighLevel}. We encode the structure of the NRN as a graph, thus \textit{NRN.root} corresponds to the root node in this graph. $NRN(x)$ is the predicted truth value of the sample $x$ given the trained NRN, $NRN$. Each node in the NRN graph has the properties, \textit{node.weights} representing the fitted weights over the logical inputs to that node, \textit{node.logic\_type} taking the value \textit{``And"}, or \textit{``Or"} depending on the type of logical node, and \textit{node.name} that is specific to Predicate nodes and represents the name supplied for that predicate. We provide examples of explanations in Figure~\ref{fig:CaliforniaExplanationExamples}.

In Algorithm~\ref{Algo:ExplainHighLevel}, $Cat$ refers to the concatenation operation for strings.  Algorithm~\ref{Algo:ExplainHighLevel} leverages Algorithm~\ref{Algo:ComputeRequiredChildValue} to compute the required value of inputs, and Algorithm~\ref{Algo:AddNot} to correctly add negation notation ``Not" to the produced explanations.

\begin{algorithm}
  \begin{algorithmic}[1]
\STATE input $logic = (NRN.node)$
\STATE input $t \in [0, 1]$
\STATE input $j$
\STATE $\textbf{i} = logic.inputs \odot \textbf{s} + (1 - logic.inputs) \odot (1 - \textbf{s})$
\STATE $c = \textbf{i}_j$
\IF{$logic.logic\_type = \text{``Or"}$}
    \STATE$c_{out} = c \times \lvert logic.weights_j \rvert$
    \STATE $s_{out} = \textbf{i}^T \lvert logic.weights \rvert - c_{out}$
    \STATE $t_{child} = (t - s_{out})\ /\ \lvert logic.weights_j \rvert$
    \STATE $t_{child} = max(0,\ min(t_{child},\ 1))$
\ENDIF
\IF{$logic.logic\_type = \text{``And"}$}
    \STATE$c_{out} = (1 - c) \times \lvert logic.weights_j \rvert$
    \STATE $s_{out} = (1 - \textbf{i})^T \lvert logic.weights \rvert - c_{out}$
    \STATE $t_{child} = (1 - t - s_{out})\ /\ \lvert logic.weights_j \rvert$
    \STATE $t_{child} = 1 - max(0,\ min(t_{child},\ 1))$
\ENDIF
\STATE $\textbf{return}\ t_{child}$
\end{algorithmic}
\caption{ComputeRequiredChildValue}
\label{Algo:ComputeRequiredChildValue}
\end{algorithm}

In algorithm Algorithm~\ref{Algo:ComputeRequiredChildValue}, the inputs include $logic$, which is a node of an NRN, $t$ which is the required output of that node as determined by the parent of this node during the recursive algorithm Algorithm~\ref{Algo:ExplainHighLevel}, and $j$ the index of the child of the node $logic$ that we are examining to determine the required value this input must have to produce at least $t$.  To determine the required value we must know what is the logic type of the parent node, either ``And", or ``Or" as we must compute the required value based on the values of the other children in this logic using an algebraic manipulation of the Modified Weighted Lukasiewicz activation function.

\begin{algorithm}
  \begin{algorithmic}[1]
    \STATE input $logic = (NRN.node)$
    \STATE input $\epsilon$
    \STATE input $negate$
    \STATE input $j$
    
    \IF{$(logic.weight_j < 0 \ and \ \textbf{not} \ negate) \ or \ (logic.weight_j \geq 0 \ and \  negate)$}
        \STATE $\epsilon \gets Cat(``NOT", ``(", \epsilon, ``)")$
    \ENDIF
    \STATE $\textbf{return} \ \epsilon$
\end{algorithmic}
\caption{AddN}
\label{Algo:AddNot}
\end{algorithm}

Algorithm~\ref{Algo:AddNot} is used to produce explanation strings that include negations.  To do so, the method takes as input $logic$ an NRN node, $\epsilon$ the current explanation string produced from the recursive algorithm Algorithm~\ref{Algo:ExplainHighLevel} thus far, $negate$ a boolean indicating if the parent of this node was negated, and $j$ the index of the input we are examining within the current logical node.  The method then determines if the negation string ``NOT" should wrap the current explanation $\epsilon$ accounting for double negation from the parent node.

\label{ExplanationGeneration}

\section{Explanation Simplification} 

Algorithm~\ref{Algo:PushingNegationsDown} describes the procedure to move all negations in the tree representing the explanation of the NRN to the leaves.  The algorithm simply traverses the tree structure in a recursive manner and adjusts the negations until negations only exist at the leaves.

\begin{algorithm*}
  \begin{algorithmic}[1]
\STATE initialize $node = Explanation.root$
\\
\STATE $\textbf{Procedure:}\ push\_negations\_down(node):$
\\
\IF {$node.logic\_type =$ ``Leaf"}
    \STATE $\textbf{return}\ $
\ELSIF{{$node.logic\_type =$ ``Not"} AND {$len(node.children) < 1$}}
    \STATE $node.convert\_not\_to\_leaf()$ \COMMENT {(Rule \# 1)}
\ELSIF{$node.logic\_type =$ ``Not"}
    \STATE $node = node.children[0]$
    \STATE $negate(node)$
\ELSE
    \FOR{$child \ \textbf{in}\ node.children$}
        \STATE $push\_negations\_down(child)$
    \ENDFOR
\ENDIF
\\
\STATE $\textbf{Procedure:}\ negate(node):$
\\
\IF{$node.logic\_type =$ ``Leaf"}
    \STATE $node.logic\_type =$ ``Not"
    \STATE $node.convert\_not\_to\_leaf()$ \COMMENT {(Rule \# 1)}
\ELSIF{{$node.logic\_type =$ ``Not"} AND {$len(node.children) < 1$} \COMMENT {(Rule \# 4)}}
    \STATE $node.logic\_type$ = 'Leaf' 
\ELSIF{$node.logic\_type =$ ``Not" \COMMENT {(Rule \# 4)}}
    \STATE $node = node.children[0]$
    \STATE $push\_negations\_down(node)$
\ELSIF{$node.logic\_type =$ ``And" \COMMENT {(Rule \# 5)}}
    \STATE $node.logic\_type =$ ``Or"
    \FOR{$child \ \textbf{in}\ node.children$}
        \STATE $negate(child)$
    \ENDFOR
\ELSIF{$node.logic\_type =$ ``Or" \COMMENT {(Rule \# 6)}} 
    \STATE $node.logic\_type =$ ``And"
    \FOR{$child \ \textbf{in}\ node.children$}
        \STATE $negate(child)$
    \ENDFOR
\ENDIF        
\end{algorithmic}
\caption{Pushing negations down}
\label{Algo:PushingNegationsDown}
\end{algorithm*}

Algorithm~\ref{Algo:CollapseConsecutiveOperands} describes the procedure collapse consecutive logical operations of the same type to a single logical operation.  This algorithm is performing the operations required for Rules 7 and 8 from Table~\ref{Table:SimplificationRules}.

\begin{algorithm*}
  \begin{algorithmic}[1]
\STATE initialize $node = Explanation.root$
\\
\STATE $\textbf{Procedure:}\ collapse\_repeated\_operands(node):$
\\
\STATE $collapse\_consecutive\_repeated\_operands\_for\_node(node)$
\FOR{$child \ \textbf{in}\ node.children$}
    \STATE $collapse\_repeated\_operands(child)$
\ENDFOR
\\
\STATE $\textbf{Procedure:}\ collapse\_consecutive\_repeated\_operands\_for\_node(node):$
\\
\STATE $same\_type\_children = list()$
\STATE $different\_type\_children = list()$
\FOR{$child \ \textbf{in}\ node.children$}
    \IF{$child.logic\_type=node.logic\_type$}
        \STATE $same\_type\_children.append(child)$
    \ELSE
        \STATE $different\_type\_children.append(child) $
    \ENDIF
\ENDFOR
\IF{$len(same\_type\_children)<1$}
    \STATE $\textbf{return}\ $
\ENDIF
\STATE $new\_children=different\_type\_children$
\FOR{$child \ \textbf{in}\ same\_type\_children$}
    \STATE $new\_children.extend(child.children)$
\ENDFOR
\STATE $node.children=new\_children$
\STATE $collapse\_consecutive\_repeated\_operands\_for\_node(node)$
\end{algorithmic}
\caption{Collapsing consecutive repeated operands}
\label{Algo:CollapseConsecutiveOperands}
\end{algorithm*}

Table~\ref{Table:SimplificationRules} shows all of the simplification rules that are used during the simplification process described above.  Each of these rules is applied to ensure that the resulting explanation is concise as possible while also representing the logic learned by the model as accurately as possible.

\begin{table*}
\caption{Explanation simplification rules}\label{Table:SimplificationRules}
\centering
\begin{tabular}{lll}
        \toprule
        \textbf{Rule} & \textbf{Input rule} & \textbf{Replacement rule} \\
        \midrule
        \multicolumn{3}{l}{\textbf{For a numerical feature $f$ and a real number $a$:}} \\
        \midrule
        1 & NOT $f < a$  & $f \geq a$ \\
        & NOT $f \leq a$  & $f > a$ \\
        & NOT $f > a$  & $f \leq a$ \\
        & NOT $f \geq a$  & $f < a$ \\
        \midrule
        2 & AND($f \leqslant a_1$, \dots, $f \leqslant a_n$) & $f \leqslant \min(a_1, \dots , a_n)$ \\
        & AND($f \geqslant a_1$, \dots , $f \geqslant a_n$) & $f \geqslant \max(a_1, \dots , a_n)$ \\
        \midrule
        3 & OR($f \leqslant a_1$, \dots , $f \leqslant a_n$) & $f \leqslant \max(a_1, \dots , a_n)$ \\
        & OR($f \geqslant a_1$, \dots , $f \geqslant a_n$) & $f \geqslant \min(a_1, \dots , a_n)$ \\
        \midrule
        \multicolumn{3}{l}{\textbf{For any predicates  $x$, $x_i$, $x_{ij}$:}} \\ 
        \midrule
        4 &  NOT NOT $x$ & $x$ \\ 
        \midrule
        5 & NOT AND ($x_1$, $x_2$, \dots , $x_n$) & OR (NOT $x_1$, NOT $x_2$, \dots , NOT $x_n$) \\
        \midrule
        6 & NOT OR ($x_1$, $x_2$, \dots , $x_n$) & AND (NOT $x_1$, NOT $x_2$, \dots , NOT $x_n$) \\
        \midrule
        7 & AND ($x_{01}$, $x_{02}$, \dots , $x_{0n_0}$, & AND ( $x_{01}$, $x_{02}$, \dots , $x_{kn_k}$) \\
         & \hspace{25pt} AND ($x_{11}$, $x_{12}$, \dots , $x_{1n_1}$) & \\
         & \hspace{25pt} AND ($x_{21}$, $x_{22}$, \dots , $x_{2n_2}$) & \\
         & \hspace{25pt} \dots & \\
         & \hspace{25pt} AND ($x_{k1}$, $x_{k2}$, \dots , $x_{kn_k}$)) & \\
        \midrule
        8 & OR ($x_{01}$, $x_{02}$, \dots , $x_{0n_0}$, & OR ( $x_{01}$, $x_{02}$, \dots , $x_{kn_k}$) \\
         & \hspace{25pt} OR ($x_{11}$, $x_{12}$, \dots , $x_{1n_1}$) & \\
         & \hspace{25pt} OR ($x_{21}$, $x_{22}$, \dots , $x_{2n_2}$) & \\
         & \hspace{25pt} \dots & \\
         & \hspace{25pt} OR ($x_{k1}$, $x_{k2}$, \dots , $x_{kn_k}$)) & \\
        \midrule
        9 & AND($x$) & $x$ \\
        & OR($x$) & $x$ \\
        \bottomrule
        \multicolumn{3}{l}{\small *Here $\leqslant$ means either $<$ or $\leq$ depending on the signs of the input rules and the limiting rule} \\
        \multicolumn{3}{l}{\small \hspace{23pt} $\geqslant$ means either $>$ or $\geq$ depending on the signs of the input rules and the limiting rule} \\
\end{tabular}
\end{table*}

\label{ExplanationSimplification}

\section{Feature Importance Generation}

The structure of the NRN is encoded as a graph, with every edge having a weight learned during training, and every leaf representing a predicate, which is mapped to a feature. A single feature can be mapped to multiple predicate nodes.  To calculate feature importance, we traverse the graph from root to leaf and aggregate the absolute values of the weights along each branch, multiplying the absolute value of the weights along the way. That is, we perform a sum-product operation along each branch over the absolute values of the weights, which results in an \textit{aggregated weight} for each leaf.  Since the same feature can be present in multiple leaves, due to the one-to-many relation with predicate nodes, we calculate the feature importance as a sum of all the \textit{aggregated weights} over leaves for which the predicate node is mapped to the corresponding feature.

\label{FeatureImportanceComp}

\section{R-NRN Hyper-Parameter Analysis}  

\begin{figure}[h]
    \includegraphics[width=1.0\linewidth]{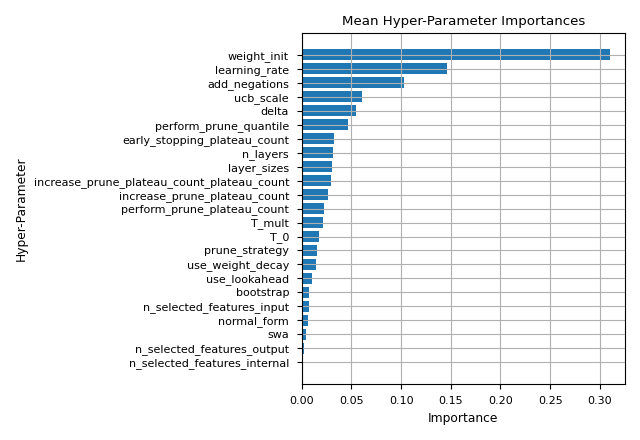}
    \caption{Mean hyper-parameter importance for R-NRN across all datasets in the benchmark. Importance values are identified by the fANOVA hyper-parameter importance evaluation algorithm.}
    \label{Fig:HyperParams}
\end{figure}

In addition to our empirical performance study, we share the results our analysis of the hyper-parameters of R-NRN to examine the importance of different aspects of the algorithm. 

Figure~\ref{Fig:HyperParams} shows mean hyper-parameter importance for R-NRN aggregated across all datasets in the benchmark. The analysis shows that R-NRN specific hyper-parameters such as \textit{add\_negations}, \textit{delta}, \textit{ucb\_scale} and others are among the most important for model performance. Weight initialization and learning rate are also important, similarly to deep learning approaches.

In addition, we find that R-NRN effectively leverages FeatureBinarizationFromTrees \cite{aix360-sept-2019} as pre-processing method for numeric data, while MLP cannot. During hyper-parameter tuning we allow the TPESampler to choose if FeatureBinarizationFromTrees is applied to numeric data. For R-NRN, all 22 experiments resulted in selecting this pre-processing method, while it was only selected in 16 experiments ($73\%$) for MLP.

\label{HyperParamAnalysis}

\section{Optimization Analysis}

\begin{figure}[h]
    \includegraphics[width=1.0\linewidth]{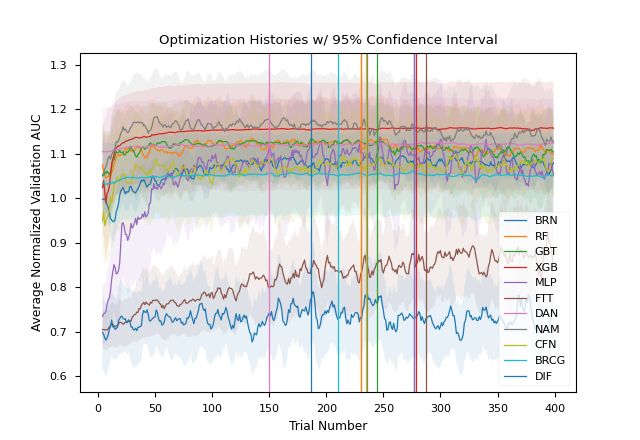}
    \caption{Mean normalized validation ROC AUC by trial across all datasets. Annotated with mean argmax of optimization history across all datasets. Vertical lines represent the average best trial index across all datasets. All trends are normalized to performance of R-NRN trial 0.}
    \label{fig:OptimizationHistory}
\end{figure}

We are interested in development of a practical algorithm for real world applications and therefore analyze the effort required to achieve optimal performance of R-NRN as compared to other algorithms. Figure~\ref{fig:OptimizationHistory} shows the mean validation performance for each algorithm across each trial in our hyper-parameter optimization. The values are normalized to the performance of the R-NRN algorithm.

From the analysis we see that while R-NRN performance improves more gradually as the search progresses, the optimal value is identified in a similar number of trials, on average, as compared with other methods. We also note that the hyper-parameter ranges for RF, XGB, and GBT are highly optimized through extensive historical use of these algorithms whereas the R-NRN algorithms search space was large.

\section{GPU Scaling}

In addition to minimizing the number of parameters, we achieve scalability with NRNs by using GPUs to broadcast tensor operations of conjunction and disjunction layers. Denote $\mathbf{A}$ and $\mathbf{V}$ as the vector operations for weighted conjunction (Equation~\ref{Equation:ModWeightedConj}) and disjunction (Equation~\ref{Equation:ModWeightedDisj}) respectively. A logical block is defined as a broadcast tensor operation with an input tensor to a block of size $(C, O, I, 1)$ where $C$ is the number of channels, $O$ is the block's output size, and $I$ is the blocks input size. The logical block contains a weight tensor of size, $(C, O, I, 1)$. Each channel $c \in C$ of a Weighted Conjunction Block ($\mathbf{A}_{block}$ Equation~\ref{Equation:Ablock}) and Weighted Disjunction Block ($\mathbf{V}_{block}$ Equation~\ref{Equation:Vblock}) contains $O$ weight vectors $\mathbf{w}_c^o \in I \times 1$, and takes as input a tiled vector $\mathbf{x}_c^o \in I \times 1$, thus producing tensors of the described shapes. In this work, channels of an NRN are leveraged for each class in a multi-class setting, though their use is dictated by the problem to be solved. An additional tensor dimension for batches of samples can be added to the operation.

\begin{gather}
    \label{Equation:Ablock}
    \begin{bmatrix}
       \mathbf{A}_0^0, \mathbf{A}_0^1, \dots, \mathbf{A}_0^O \\
       \mathbf{A}_1^0, \mathbf{A}_1^1, \dots, \mathbf{A}_1^O \\
       \vdots \\
       \mathbf{A}_C^0, \mathbf{A}_C^1, \dots, \mathbf{A}_C^O \\
    \end{bmatrix}
\end{gather}

\begin{gather}
    \label{Equation:Vblock}
    \begin{bmatrix}
       \mathbf{V}_0^0, \mathbf{V}_0^1, \dots, \mathbf{V}_0^O \\
       \mathbf{V}_1^0, \mathbf{V}_1^1, \dots, \mathbf{V}_1^O \\
       \vdots \\
       \mathbf{V}_C^0, \mathbf{V}_C^1, \dots, \mathbf{V}_C^O \\
    \end{bmatrix}
\end{gather}

\label{gpu_scaling}

\section{R-NRN Algorithm Details}

Algorithm~\ref{Algo:BRNOverview} gives the high level procedure for the R-NRN algorithm.  $\mathcal{T}$ is the number of epochs to train for.  $M_1(\Phi)$ is Model-1 in the R-NRN algorithm.  $M_2(\Theta)$ is Model-2 in the R-NRN algorithm.  $\alpha$ is a learning rate to use for Gradient Descent.  $L$ is a loss function such as Binary Cross Entropy.  $R$ is one of the Reward Strategies described in the next section.  $UpdatePolicy$ is the policy update method for BAMB described in Algorithm~\ref{Algo:BRNSampling}.

\begin{algorithm}
    \centering
    \caption{R-NRN}\label{Algo:BRNOverview}
    \begin{algorithmic}[1]
        \STATE initialize $\mathcal{T}\text{,} \mathcal{T} \in \mathbb{Z}\text{,} \mathcal{T} > 0$
        \STATE initialize $M_2(\Theta)$
        \STATE initialize $M_1(\Phi)$
        \STATE initialize $\alpha, \alpha \in \mathcal{R}, \alpha > 0$
        
        \FOR{$t\in\mathcal T$}
            \STATE $\hat{y} = M_2(\Theta)$
            \STATE $l = L(\hat{y}, y)$
            \STATE $\Theta \gets \Theta - \alpha\nabla M_2(\Theta)$
            \IF{update criteria met}
                \STATE $r = R(M_2(\Theta))$
                \STATE $\Phi \gets \text{UpdatePolicy}(\Phi, r)$
            \ELSIF{sampling criteria met}
                \STATE $M_2(\Theta) \gets M_1(\Phi)$
            \ENDIF
        \ENDFOR
    \end{algorithmic}
\end{algorithm}

Algorithm~\ref{Algo:BRN} shows a more detailed version of the R-NRN logic induction algorithm detailing the interaction of $\kappa$, $\iota$, and $\tau$, described in the main body of our paper, to determine when pruning and growing of the NRN.  We also show $\rho$ the threshold for pruning and producing rewards with one of the reward strategies, $pc$ which is a variable to record the current number of epochs that performance has plateaued for, $pt$.  This algorithm also shows $\delta$, described in the main body of the paper used in Algorithm~\ref{Algo:BRNSampling}.

\begin{algorithm}[h]
    \centering
    \caption{R-NRN detailed}\label{Algo:BRN}
    \begin{algorithmic}[1]
        \STATE initialize $\mathcal{T}\text{,} \mathcal{T} \in \mathbb{Z}\text{,} \mathcal{T} > 0$
    \STATE initialize $M_2(\Theta)$
    \STATE initialize $M_1(\Phi)$
    \STATE initialize $\rho \in [0, 1]$
    \STATE initialize $\delta \in (0, \infty]$
    \STATE initialize $\alpha \in \mathcal{R}, \alpha > 0$
    \STATE initialize $\text{pc} \in [0, \infty) $
    \STATE initialize $\tau \in [0, \infty) $
    \STATE initialize $\iota \in [0, \infty) $
    \STATE initialize $\kappa \in [0, \infty) $
    \STATE initialize $l_{min} = \infty$
    \FOR{$t\in\mathcal T$}    
        \STATE $\hat{y} = M_2(\Theta)$
        \STATE $l = L(\hat{y}, y)$
        \STATE $\Theta \gets \Theta - \alpha\nabla M_2(\Theta)$
        \IF{$l < l_{min}$}
            \STATE $l_{min} \gets l$
            \STATE $\text{pc} \gets 0$
            \STATE $\mathbf{r} = \text{BmabReward}(\Theta, \rho)$
            \STATE $\Phi \gets \text{BmabPolicyUpdate}(\mathbf{r})$
        \ELSE
            \STATE $\text{pc} \gets \text{pc} + 1$
            \IF{$\text{pc} > \kappa$}
                \STATE $M_2(\Theta) \gets \text{R-NRNSampling}(\rho, \delta, \Phi)$
                \STATE $\text{pc} \gets 0$
            \ELSIF{$\text{pc} > \tau$}
                \STATE $\kappa \gets \kappa + \iota$
            \ENDIF
        \ENDIF
    \ENDFOR
    \end{algorithmic}
\end{algorithm}

Algorithm~\ref{Algo:BRNSampling} shows the process to sample new predicates while training R-NRN.  $\hat{I}$ is an empty set that will be used to store the predicates that should be sampled.  $a$ and $b$ are the lower and upper bounds that will be used to draw from a uniform distribution when sampling new predicate weights.  $\hat{\Phi}$ is initialized to the weights from Model-1, $M_1(\Phi)$.  $\mathcal{I}$ is initialized to the number of features in the data $D^{n x m}$.  $\mathcal{K}$ is initialized to the number of predicates contained within the logic of the first layer of the NRN Model-2 weights, $\theta$.  The procedure is used to sample new predicates and place them into the Model-2 NRN as described in the main body of our work.

\begin{algorithm}[h]
    \centering
    \caption{R-NRNSampling}\label{Algo:BRNSampling}
    \begin{algorithmic}[1]
        \STATE $D^{n \times m}$
        \STATE initialize $\hat{I} = \emptyset$
        \STATE initialize $a \in \mathcal{R}$
        \STATE initialize $b \in \mathcal{R}$
        \STATE initialize $\hat{\Phi} = \Phi$
        \STATE initialize $\mathcal{I} = \mathcal{Z} \in (0, m]$
        \STATE initialize $\mathcal{K} = \mathcal{Z} \in (0, \lvert \Theta_{0} \rvert]$
        \STATE $\mathcal{M} = f: k \mapsto i, \forall \ k \in \mathcal{K}, , \forall \ i \in 
        \mathcal{I}$
        \FOR{$k \in \mathcal{K}$}
            \IF{$abs(\Theta_0^k) > \text{percentile}(abs(\Theta_0), \rho)$}
                \STATE $i = \mathcal{M}(k)$
                \STATE $\hat{\Phi}_{i} \gets \frac{\Phi{i}}{\delta}$
                \STATE $\hat{I} \gets \hat{I} \cup \{i\}$
            \ENDIF
        \ENDFOR
        \FOR{$k \in \mathcal{K}$}
            \IF{$abs(\Theta_0^k) \leq \text{percentile}(abs(\Theta_0), \rho)$}
                \STATE $\hat{i} \sim \hat{\Phi}$
                \IF{$\hat{i} \in \hat{I}$}
                    \STATE $\mathbf{k} = \mathbf{I}_{\mathcal{M}(k) = \hat{i}, \forall k \in \mathcal{K}}$
                    \STATE $\mathbf{u} \sim Uniform(a, b), \lvert u \rvert = \lvert K \rvert$
                    \STATE $\Theta_0^k = -1 \times sgn(\sum_{j\in\mathcal{K}}[\mathbf{k} \odot \Theta_0]_j) \times \mathbf{u}_k$
                \ENDIF
            \ENDIF
        \ENDFOR
    \end{algorithmic}
\end{algorithm}

\section{Reward strategies}
\label{RewardStrategies}

\begin{algorithm}[h]
    \centering
    \caption{Class reward}\label{Algo:ClassReward}
    \begin{algorithmic}[1]
        \STATE $D^{n \times m}$
        \STATE initialize $\mathbf{r} \in m = 0$
        \STATE initialize $\mathcal{I} = \mathcal{Z} \in (0, m]$
        \STATE initialize $\mathcal{K} = \mathcal{Z} \in (0, \lvert \Theta_{0} \rvert]$
        \STATE $\mathcal{M} = f: k \mapsto i, \forall \ k \in \mathcal{K}, , \forall \ i \in 
        \mathcal{I}$
        \FOR{$k \in \mathcal{K}$}
            \IF{$abs(\Theta_0^k) > \text{percentile}(abs(\Theta_0), \rho)$}
                \STATE $\mathbf{r}_{i = \mathcal{M}(k)} = \mathbf{r}_{i = \mathcal{M}(k)} + abs(\Theta_0^k)$
            \ENDIF
        \ENDFOR
        \STATE $\textbf{return r}$
    \end{algorithmic}
\end{algorithm}

Algorithm~\ref{Algo:ClassReward} describes the class reward method detailed in the main body of our paper.  In Algorithm~\ref{Algo:LogicReward}, we collect the outputs of each logic in the first layer of the Model-2 NRN. Those values are used to evaluate each logic's predictive performance individually using any evaluation criteria, such as ROC AUC or Average Precision. Rewards then correspond to the produced evaluation metric for each logic and are assigned to each Predicate that is an input to that logical node.

\begin{algorithm}[h]
  \begin{algorithmic}[1]
    \STATE $D^{n \times m}$
    \STATE initialize $\mathbf{r} \in m = 0$
    \STATE initialize $\mathcal{I} = \mathcal{Z} \in (0, m]$
    \STATE initialize $\mathcal{K} = \mathcal{Z} \in (0, \lvert \Theta_{1} \rvert]$
    \STATE $\mathcal{M} = f: k \mapsto i, \forall \ k \in \mathcal{K}, , \forall \ i \in 
    \mathcal{I}$
    \STATE initialize $\textbf{a} = 0$
    \FOR{$k \in \mathcal{K}$}
        \STATE $\textbf{a}_k = roc\_auc\_score(\Theta_0(x)^k)$
    \ENDFOR
    \FOR{$k \in \mathcal{K}$}
        \IF{$\textbf{a}_k > \text{percentile}(\textbf{a}, \rho)$}
            \STATE $\mathbf{r}_{i = \mathcal{M}(k)} = \mathbf{r}_{i = \mathcal{M}(k)} + \textbf{a}_k$
        \ENDIF
    \ENDFOR
    \STATE $\textbf{return r}$
    \end{algorithmic}
    \caption{BMAB reward - logic}
    \label{Algo:LogicReward}
\end{algorithm}

While Algorithm~\ref{Algo:LogicReward} can produce logic level rewards for the BMAB, it is slow to compute. In Algorithm~\ref{Algo:LogicClassReward}, we collect the weights corresponding to each logic in the first layer of the Model-2 NRN and use those directly as rewards. The approach is fast to compute since we are only required to examine the state of the Neural Reasoning Network, and produces a reward at the logic level rather than the predicate level, similarly to Algorithm~\ref{Algo:LogicReward}.  In all algorithms $D^{n x m}$ is the data used to train the algorithm which is used to initialize the Reward strategy based on the shap of the data.  $\Theta$ represents the Model-2 NRN weights, where $\Theta_0$ represents the weights for the first layer in the NRN and $\Theta_1$ represents the weights for the second layer in the NRN.

\begin{algorithm}[h]
  \begin{algorithmic}[1]
    \STATE $D^{n \times m}$
    \STATE initialize $\mathbf{r} \in m = 0$
    \STATE initialize $\mathcal{I} = \mathcal{Z} \in (0, m]$
    \STATE initialize $\mathcal{K} = \mathcal{Z} \in (0, \lvert \Theta_{1} \rvert]$
    \STATE $\mathcal{M} = f: k \mapsto i, \forall \ k \in \mathcal{K}, , \forall \ i \in 
    \mathcal{I}$
    \FOR{$k \in \mathcal{K}$}
        \IF{$abs(\Theta_1^k) > \text{percentile}(abs(\Theta_1), \rho)$}
            \STATE $\mathbf{r}_{i = \mathcal{M}(k)} = \mathbf{r}_{i = \mathcal{M}(k)} + abs(\Theta_1^k)$
        \ENDIF
    \ENDFOR
    \STATE $\textbf{return r}$
    \end{algorithmic}
    \caption{BMAB reward - logic class}
    \label{Algo:LogicClassReward}
\end{algorithm}

\section{Hyper-Parameter Ranges}

The following tables contain the hyper-parameter ranges used in our experiments for each algorithm.  We use the hyper-parameter ranges from \cite{grinsztajn2022why} for RF, XGB, GBT.  For NAM, CFN, FTT, BRCG, DAN, and DIF we use hyper-parameter ranges suggested by the authors in their publications or we infer them from the author's code bases.  For MLP we use many of the hyper-parameter ranges described in \cite{DBLP:journals/corr/abs-2106-11189} but we do not use data augmentation in our experiments and therefore exclude those hyper-parameters.

\begin{table}[!h]
    \centering
    \begin{tabular}{ll}
        \toprule
        Parameter & Range \\
        \midrule
        patience & UniformInt[50, 200] \\
        lr & Uniform[0.008, 0.02] \\
        layer & [8, 20, 32] \\
        base\_outdim & [64, 96] \\
        k & [5, 8] \\
        drop\_rate & [0.0, 0.1] \\
        \bottomrule
    \end{tabular}
    \caption{DANet hyper-parameter ranges}
    \label{Table8}
\end{table}

\begin{table}[!h]
    \centering
    \begin{tabular}{ll}
        \toprule
        Parameter & Range \\
        \midrule
        num\_neurons & UniformInt[number of features, \\
        & (number of features) * 4] \\
        num\_layers & UniformInt[2, 8] \\
        tau & [1, 1/0.3, 1/0.1, 1/0.03, 1/0.01] \\
        learning\_rate & Uniform[0.001, 0.1] \\
        training\_bit\_count & [16, 32, 64] \\
        connections & [``unique", ``random"] \\
        grad\_factor & Uniform[0.001, 0.5] \\
        \bottomrule
    \end{tabular}
    \caption{DiffLogic hyper-parameter ranges}
    \label{Table11}
\end{table}

\begin{table}[!h]
    \centering
    \begin{tabular}{ll}
        \toprule
        Parameter & Range \\
        \midrule
        layer\_sizes & UniformInt[2, 30] \\
        n\_layers & UniformInt[1, 6] \\
        n\_selected\_features\_input & UniformInt[2, 12] \\
        n\_selected\_features\_internal & UniformInt[2, 10] \\
        n\_selected\_features\_output & UniformInt[2, 10] \\
        perform\_prune\_quantile & Uniform[0.05, 0.9] \\
        perform\_prune\_pc ($\kappa$) & UniformInt[1, 8] \\
        ip\_pc ($\iota$) & UniformInt[0, 20] \\
        ip\_plateau\_count\_pc ($\tau$) & UniformInt[10, 30] \\
        ucb\_scale & Uniform[1.0, 2.0] \\
        normal\_form & [cnf, dnf] \\
        prune\_strategy & [class, logic, logic\_class] \\
        delta & Uniform[1.0, 12.0] \\
        bootstrap & [True, False] \\
        weight\_init & Uniform[0.01, 1.0] \\
        add\_negations & [True, False] \\
        learning\_rate & Uniform[0.0001, 0.15] \\
        early\_stopping\_plateau\_count & UniformInt[25, 50] \\
        t\_0 & UniformInt[2, 10] \\
        t\_mult & UniformInt[1, 5] \\
        use\_swa & [True, False] \\
        \midrule
        use\_l1 & [True, False] \\
        l1\_lambda & Uniform[0.00001, 0.1] \\
        \midrule
        use\_weight\_decay & [True, False] \\
        weight\_decay\_alpha & Uniform[0.00001, 0.1] \\
        \midrule
        use\_lookahead & [True, False] \\
        lookahead\_steps & UniformInt[4, 15] \\
        lookahead\_steps\_size & Uniform[0.5, 0.8] \\
        \midrule
        fbft\_tree\_num & UniformInt[2, 20] \\
        fbft\_tree\_depth & UniformInt[2, 10] \\
        fbft\_feature\_selection & Uniform[0.3, 1.0] \\
        fbft\_thresh\_round & UniformInt[2, 6] \\
        \bottomrule
    \end{tabular}
    \caption{R-NRN hyper-parameter ranges. fbft\_ refers to features for feature\_binarization\_from\_trees. ip\_ refers to increase\_prune.  pc\_ refers to plateau\_count}
    \label{Table6}
\end{table}

\begin{table}[!h]
    \centering
    \begin{tabular}{ll}
        \toprule
        Parameter & Range \\
        \midrule
        network\_depth & UniformInt[2, 50] \\
        variant & [``diagonalized", \\
        & ``ladder\_of\_ladders", \\
        & ``dlolc"] \\
        lr & Uniform[0.0001, 0.01] \\
        momentum & Uniform[0.85, 0.95] \\
        early\_stopping\_plateau\_count & UniformInt[20, 200] \\
        weight\_decay & Uniform[.00001, 0.1] \\
        \bottomrule
    \end{tabular}
    \caption{CoFrNet hyper-parameter ranges. dlolc refers to diag\_ladder\_of\_ladder\_combined}
    \label{Table10}
\end{table}

\begin{table}[!h]
    \centering
    \begin{tabular}{ll}
        \toprule
        Parameter & Range \\
        \midrule
        attn\_dropout & Uniform[0.0, 0.5] \\
        ff\_dropout & Uniform[0.0, 0.5] \\
        depth & [5, 6, 7] \\
        heads & [7, 8, 9] \\
        learning\_rate & Uniform[0.001, 0.1] \\
        early\_stopping\_plateau\_count & UniformInt[25, 50] \\
        t\_0 & UniformInt[2, 10] \\
        t\_mult & UniformInt[1, 5] \\
        \midrule
        use\_weight\_decay & [True, False] \\
        weight\_decay & Uniform[0.00001, 0.1] \\
        \midrule
        use\_lookahead & [True, False] \\
        lookahead\_steps & UniformInt[4, 15] \\
        lookahead\_steps\_size & Uniform[0.5, 0.8] \\
        \bottomrule
    \end{tabular}
    \caption{FT-Transformer hyper-parameter ranges}
    \label{Table4}
\end{table}

\begin{table}[!h]
    \centering
    \begin{tabular}{ll}
        \toprule
        Parameter & Range \\
        \midrule
        lambda0 & Uniform[.0001, 0.01] \\
        lambda1 & Uniform[.0001, 0.01] \\
        cnf & [True, False] \\
        k & UniformInt[5, 15] \\
        d & UniformInt[5, 15] \\
        b & UniformInt[5, 15] \\
        fbft\_tree\_num & UniformInt[2, 20] \\
        fbft\_tree\_depth & UniformInt[2, 10] \\
        fbft\_feature\_selection & Uniform[0.3, 1.0] \\
        fbft\_thresh\_round & UniformInt[2, 6] \\
        iter\_max & 100 \\
        time\_max & 120 \\
        solver & ``ECOS" \\
        \bottomrule
    \end{tabular}
    \caption{BRCG hyper-parameter ranges. fbft refers to features for feature\_binarization\_from\_trees.}
    \label{Table9}
\end{table}

\begin{table}[!h]
    \centering
    \begin{tabular}{ll}
        \toprule
        Parameter & Range \\
        \midrule
        layer\_sizes & [(10,), (20,), (50,), (10, 10), \\
        & (20, 20), (50, 50)] \\
        activation & [logistic, tanh, relu] \\
        learning\_rate & Uniform[0.001, 0.1] \\
        early\_stopping\_plateau\_count & UniformInt[25, 50] \\
        t\_0 & UniformInt[2, 10] \\
        t\_mult & UniformInt[1, 5] \\
        use\_swa & [True, False] \\
        use\_batchnorm & [True, False] \\
        \midrule
        use\_dropout & [True, False] \\
        dropout\_percent & Uniform[0.0, 0.8] \\
        \midrule
        use\_weight\_decay & [True, False] \\
        weight\_decay & Uniform[0.00001, 0.1] \\
        \midrule
        use\_lookahead & [True, False] \\
        lookahead\_steps & UniformInt[4, 15] \\
        lookahead\_steps\_size & Uniform[0.5, 0.8] \\
        \midrule
        fbft\_tree\_num & UniformInt[2, 20] \\
        fbft\_tree\_depth & UniformInt[2, 10] \\
        fbft\_feature\_selection & Uniform[0.3, 1.0] \\
        fbft\_thresh\_round & UniformInt[2, 6] \\
        \bottomrule
    \end{tabular}
    \caption{MLP hyper-parameter ranges.  fbft refers to features for feature\_binarization\_from\_trees.}
    \label{Table3}
\end{table}

\begin{table}[!h]
    \centering
    \begin{tabular}{ll}
        \toprule
        Parameter & Range \\
        \midrule
        max\_depth & UniformInt[1, 11] \\
        n\_estimators & [100, 200, 6000] \\
        min\_child\_weight & LogUniform(1, 100) \\
        subsample & Uniform[0.5, 1.0] \\
        eta & LogUniform[0.00001, 0.1] \\
        colsample\_bylevel & Uniform[0.5, 1.0] \\
        colsample\_bytree & Uniform[0.5, 1.0] \\
        gamma & LogUniform[0.00000001, 7.0] \\
        reg\_lambda & LogUniform[1.0, 4.0] \\
        reg\_alpha & LogUniform[0.00000001, 100.0] \\
        \bottomrule
    \end{tabular}
    \caption{XGBoost hyper-parameter ranges}
    \label{Table5}
\end{table}

\begin{table}[!h]
    \centering
    \begin{tabular}{ll}
        \toprule
        Parameter & Range \\
        \midrule
        criterion & [gini, entropy] \\
        n\_estimators & LogUniformInt[10, 3000] \\
        max\_depth & [None, 2, 3, 4] \\
        max\_features & [``sqrt", ``sqrt", \\ 
         & ``log2", None, 0.1, 0.2, 0.3, 0.4, \\
         & 0.5, 0.6, 0.7, 0.8, 0.9] \\
        min\_samples\_split & [2, 3] \\
        min\_samples\_leaf & LogUniformInt[2, 50] \\
        bootstrap & [True, False] \\
        min\_impurity\_decrease & [0.0, 0.01, 0.02, 0.05] \\
        \bottomrule
    \end{tabular}
    \caption{Random forest hyper-parameter ranges.}
    \label{Table2}
\end{table}

\begin{table}[!ht]
    \centering
    \begin{tabular}{ll}
        \toprule
        Parameter & Range \\
        \midrule
        learning\_rate & Uniform[0.001, 0.1] \\
        output\_regularization & Uniform[0.001, 0.1] \\
        l2\_regularization & Uniform[0.000001, 0.0001] \\
        dropout & [0, 0.05, 0.1, 0.2, 0.3, 0.4, \\
        & 0.5, 0.6, 0.7, 0.8, 0.9] \\
        feature\_dropout & [0, 0.05, 0.1, 0.2] \\
        training\_epochs & 1000 \\
        batch\_size & 1024 \\
        decay\_rate & 0.995 \\
        num\_basis\_functions & 1000 \\
        units\_multiplier & 2 \\
        shallow & False \\
        early\_stopping\_epochs & 60 \\
        \bottomrule
    \end{tabular}
    \caption{Neural Additive Models hyper-parameter ranges}
    \label{Table7}
\end{table}

\end{document}